\documentclass[a4paper,fleqn]{cas-sc}

\usepackage[numbers]{natbib}
\usepackage{rotating}
\usepackage{float}
\usepackage{hyperref}
\hypersetup{
    colorlinks=true,
    linkcolor=blue,
    citecolor=blue,
    urlcolor=blue
}
\usepackage{xcolor}
\usepackage{colortbl}
\usepackage{array}
\usepackage{algorithm}
\usepackage{algpseudocode}  % Using algpseudocode for better formatting
\usepackage{algorithmicx}

\def\tsc#1{\csdef{#1}{\textsc{\lowercase{#1}}\xspace}}
\tsc{WGM}
\tsc{QE}
\tsc{EP}
\tsc{PMS}
\tsc{BEC}
\tsc{DE}
\begin{document}
\let\WriteBookmarks\relax
\def\floatpagepagefraction{1}
\def\textpagefraction{.001}

\shorttitle{Mitigating LLM Hallucinations in Healthcare}

\shortauthors{Zeba et~al.}

\title [mode = title]{Mitigating Hallucinations in Healthcare LLMs with Granular Fact-Checking
and Domain-Specific Adaptation}                      
% Authors
% =========================

\author[1,2]{Musarrat Zeba}
\author[1,2]{Abdullah Al Mamun}
\author[1,3]{Kishoar Jahan Tithee}
\author[1,2]{Debopom Sutradhar}
\author[4]{Mohaimenul Azam Khan Raiaan}
[orcid=0009-0006-4793-5382]
\cormark[1]
\ead{mohaimenul.raiaan@monash.edu}
\author[5]{Saddam Mukta}
\author[6]{Reem E. Mohamed}
\author[7]{Md Rafiqul Islam}
\author[7]{Yakub Sebastian}
\author[6]{Mukhtar Hussain}
\author[7]{Sami Azam} [orcid=0000-0001-7572-9750]
\cormark[1]
\ead{sami.azam@cdu.edu.au}
% =========================
% Affiliations
% =========================

\affiliation[1]{organization={Applied Artificial Intelligence and Intelligent Systems (AAIINS) Laboratory},
    city={Dhaka},
    postcode={1217},
    country={Bangladesh}
}

\affiliation[2]{organization={Department of Computer Science and Engineering, United International University},
    city={Dhaka},
    postcode={1212},
    country={Bangladesh}
}

\affiliation[3]{organization={Department of Computer Science and Engineering, Daffodil International University},
    city={Dhaka},
    postcode={1341},
    country={Bangladesh}
}

\affiliation[4]{organization={Department of Data Science and Artificial Intelligence, Monash University},
    city={Clayton},
    state={VIC},
    postcode={3153},
    country={Australia}
}

\affiliation[5]{organization={Department of Software Engineering, Lappeenranta-Lahti University of Technology},
    city={Lappeenranta},
    postcode={53850},
    country={Finland}
}

\affiliation[6]{organization={Faculty of Science and Information Technology, Charles Darwin University},
    city={Sydney},
    state={NSW},
    country={Australia}
}

\affiliation[7]{organization={Faculty of Science and Technology, Charles Darwin University},
    city={Casuarina},
    state={NT},
    postcode={0909},
    country={Australia}
}


\cortext[cor1]{Corresponding authors}

\begin{abstract}
In healthcare, it is essential for any Large Language Model (LLM)-generated output to be reliable and accurate, particularly in cases involving decision-making and patient safety. However, the outputs are often unreliable in such critical areas due to the risk of hallucinated outputs from the LLMs. To address this issue, we propose a fact-checking module that operates independently of any LLM, along with a domain-specific summarization model designed to minimize hallucination rates. Our model is fine-tuned using Low-Rank Adaptation ({LoRA}) on the MIMIC-III dataset and is paired with the fact-checking module, which uses numerical tests for correctness and logical checks at a granular level through discrete logic in natural language processing (NLP) to validate facts against electronic health records (EHRs). We trained the LLM on the full MIMIC-III dataset. For evaluation of the fact-checking module, we sampled 104 summaries, extracted them into 3,786 propositions, and used these as facts. The fact-checking module achieves a precision of 0.8904, a recall of 0.8234, and an F1-score of 0.8556. Additionally, the LLM summary achieves a ROUGE-1 score of 0.5797 and a BERTScore of 0.9120 for summary quality.
\end{abstract}

% \includegraphics{figs/grabs.pdf}


% Keywords
\begin{keywords}
large language models \sep hallucination  mitigation \sep clinical text summarization \sep fact-checking \sep domain-specific adaptation
\end{keywords}

\maketitle

\section{Introduction}

The medical sector is rapidly adopting Artificial Intelligence (AI) nowadays, but there are issues regarding the reliability of the outputs in the real-world use case \cite{LIN2025100868, han2025dynamic}. Large language models (LLMs) usually have great contributions in healthcare when used \cite{xu2025staf}. However, hallucinations are a major drawback in the use of LLMs in these sectors \cite{kim2025medical}. Because, as a safety-critical domain, healthcare can not tolerate diagnostic or factual errors \cite{li2025taming}. Therefore, critical areas are often not encouraged to use and depend on these AI tools. Studies reveal frequent factual errors in LLM-generated clinical summaries, and patient reports. These errors further reduce clinician trust and slow adoption in practice \cite{kim2025medical}.

Hallucination continues to be a fundamental challenge despite the rapid advancement of LLMs \cite{rawte2023survey,maynez2020faithfulness}. This phenomenon involves the confident creation of information that is either unverified or completely false \cite{heo2025halucheck}. It represents a significant hurdle for the use of LLMs in clinical practice \cite{pal2023med}. In the healthcare domain, minor inaccuracies such as incorrect dosage of medications, invented diagnoses, or distorted laboratory values can have serious consequences for patient safety and disrupt clinical workflows that can drastically erode trust in AI-assisted decision-making \cite{geroimenko2025generative, lv2025mire}.

The exciting potential and challenges of large language models (LLMs) in healthcare are now coming to light through recent research \cite{lv2025mire, karabacak2023embracing, wu2025contrastive}. For instance, Med-PaLM 2 has shown strong reasoning abilities in clinical question answering. They have demonstrated exceptional performance and have showcased how specialized models can help with reasoning in complex clinical situations \cite{singhal2023clinicalknowledge}. To improve both the contextual accuracy and the readability of medical texts, ClinicalGPT has also been introduced. It adapts general LLMs to create radiology and discharge summaries \cite{wang2023clinicalgpt}. Similarly, some efforts in medical summarization by Tang et al. \cite{Tang2023LLMSummarization}, Xu et al. \cite{xu2024opportunities}, and Lin et al. \cite{LIN2025100868} demonstrate the utility of LLMs in condensing lengthy clinical notes into concise narratives that aid communication between healthcare providers. Hallucination detection and factuality verification frameworks, including CHECK \cite{garcia2025check}, retrieval-augmented generation methods \cite{lewis2020retrievalaugmentedgeneration, kong2025halugnn, shuster2021retrieval}, graph based methods \cite{kong2025halugnn, chen2025dairygoatqa}, multi-agent approach \cite{shi2025mitigating} and survey analyzes \cite{ji2023survey, Huang_2025}, highlight ongoing efforts to reduce errors and improve reliability, which eventually extend their focus beyond mere summarization. Therefore, the key question is not whether LLMs can write convincing medical text, but whether clinicians and researchers can place their trust in the results these models provide.

Despite these advances, some crucial limitations remain. Studies consistently show that even leading-edge LLMs hallucinate in 2–5\% of the generated medical summaries, with ingenious inaccuracies often escaping the notice of the clinician \cite{Tang2023LLMSummarization, xu2024opportunities}. A widely cited investigation reported that more than 40\% of the summaries generated by an LLM contained factual errors, ranging from incorrect prescriptions to fabricated diagnoses \cite{Tang2023LLMSummarization}.

In this work, we address these limitations by introducing an LLM-free fact-checking system for clinical text verification. Unlike previous approaches, our pipeline eliminates LLM dependency during the evaluation stage, replacing it with a deterministic and transparent mechanism that combines several propositional logical consistency checks together to work as a whole. In parallel, we fine-tuned a domain-specialized generator on more than 40,000 patient records from the MIMIC-III dataset \cite{johnson2016mimic}, from which 26,104 discharge summaries were extracted for training and evaluation, using LoRA \cite{hu2022lora}, a parameter-efficient fine-tuning technique that injects lightweight rank-decomposition matrices into pre-trained weights. This approach enables effective domain adaptation while significantly reducing computational overhead, resulting in clinically grounded summaries with fewer hallucinations at the generation stage.

As the framework is developed using the MIMIC-III dataset \cite{johnson2016mimic}, which consists of large-scale clinical data covering a broad range of patient conditions, treatments, and clinical observations, it supports learning from diverse medical scenarios. The approach focuses on structured representations of clinical information, which enables it to operate across varied medical contexts. For clarity of presentation, selected examples are included throughout the paper to illustrate the behavior of the system at the proposition level.

\begin{figure}[]
    \centering
    \includegraphics[scale=0.18]{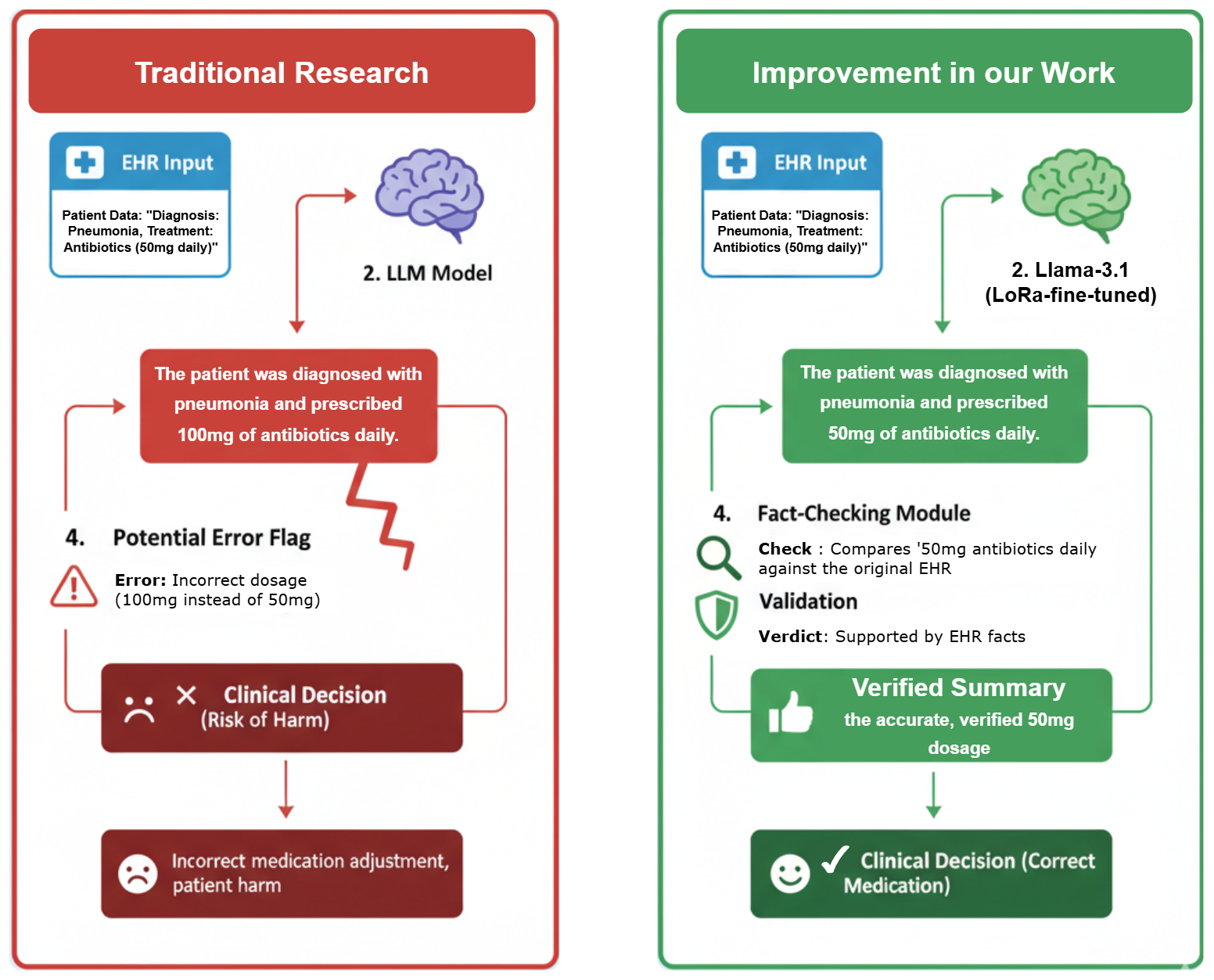}
    \caption{\textbf{The figure shows a comparison between traditional LLM-only summarization in healthcare and our proposed LLM with the integration of the Fact-checking module. The left side shows that the conventional LLMs may produce hallucinated outputs, which can be factually incorrect (e.g., incorrect medication dosage is wrong). These errors can lead to risky medical decisions. On the other hand, the right side shows the illustration of the benefit of using a fine-tuned specific LLM with the fact-checking module that verifies each claim against the EHR data.}}
    \label{fig:before_after_factcheck}
    \vspace{-15pt}
\end{figure}

Clinical decision-making is heavily based on accurate summaries of Electronic Health Records (EHRs). Traditional language models generate these summaries, but often introduce errors, such as incorrect dosages or diagnoses. These errors can be risky to patients. Our approach solves this problem by combining a specialized LLaMA model with a fact-checking module. Together, these two combine to minimize hallucinations as well as flag the ones that yet remain. Figure \ref{fig:before_after_factcheck} shows how, in the traditional method, errors, such as incorrect dosages, can lead to harmful clinical decisions. In contrast, our method ensures that every fact is validated against the original EHR. This allows clinicians to rely on LLM-generated summaries and outcomes for their accurate results. 

The main contributions of this work are as follows.

\begin{itemize}
    \item Introduces an LLMs-free Fact-Checking module that applies discrete logic to evaluate negation, implication, temporal consistency, numerical checks, and cosine similarity for assigning verdicts, systematically verifying each proposition claim in a summary against the corresponding patient’s EHRs with transparency and independence from additional LLMs.
    \item Proposes a deterministic verification mechanism that breaks down both generated summaries and EHRs into atomic propositions, allowing one-to-one comparisons for independent fact-checking at a granular level.
    
    \item Develops a domain-specific summarization model by fine-tuning LLaMA-3.1-8B with Low-Rank Adaptation (LoRA) on 26,104 MIMIC-III discharge summaries, enabling efficient adaptation and generating clinically accurate summaries with reduced hallucination rates.
    
    % \item Achieves robust performance on 104 MIMIC-III summaries with 3,786 propositions, attaining ROUGE-1 of 0.5797, BERTScore of 0.9120 for summarization quality, and fact-checking F1-score of 0.8556, with external medical professional validation confirming 85\% clinical relevance, surpassing existing study benchmarks.
\end{itemize} 
\section{Related Work}
Over the last two decades, Natural Language Processing (NLP) and LLMs have advanced from rule-based systems to neural architectures capable of generating fluent and contextually coherent text. Despite these advances, their use in healthcare is limited due to concerns about reliability, interpretability, and ethical issues, with hallucination being a major barrier to adoption. In this section, we present a review of recent studies on hallucination detection, fact verification, and error correction in clinical NLP and highlight approaches such as automated fact-checking with natural language inference (NLI), LLM-based and LLM-free verification pipelines, error mitigation strategies, and proposition-level consistency checking.

\subsection{Hallucination in Medical Large Language Models}
LLMs have achieved great success in various natural language tasks. This improvement helps the healthcare industry by providing opportunities, such as clinical decision making and automated summary processing of patient records \cite{yang2023large}. However, the implications are limited in the healthcare sector due to the tendency to generate hallucinated or unsupported statements \cite{zhao2024fact, ong2024ethical}. Several studies have focused on detecting or mitigating such factual inconsistencies.  

Joseph et al. \cite{joseph2024factpico} introduced the FACTPICO benchmark, a framework aimed at evaluating how well medical summaries generated by LLMs like GPT-4, LLaMA-2, and Alpaca align with factual information using the PICO (Population, Intervention, Comparator, Outcome) structure. Their results showed a trade-off between factual accuracy and linguistic fluency: Alpaca was more accurate but less coherent, while LLaMA-2 and GPT-4 were more fluent but prone to errors, with hallucination rates of LLaMA-2 reaching 38\%. While the PICO-based evaluation aligned better with clinician judgments than generic automatic metrics, the reliance on a small set of high-level PICO elements introduces important limitations. Many clinically relevant hallucinations, such as misstated temporal qualifiers, incorrect dosages, or omitted comorbidities, may not alter the P, I, C, or O labels and therefore remain invisible to the benchmark. Moreover, mapping long free-text summaries into discrete PICO fields is inherently lossy and can collapse multiple distinct factual propositions into a single category, making it difficult to assess whether each statement in the summary is grounded in the underlying evidence. Consequently, FACTPICO is more suitable for coarse-grained evaluation of summary fidelity than for instance-level inconsistency detection, which motivates finer-grained, proposition-level fact-checking approaches.

% Joseph et al. \cite{joseph2024factpico} introduced the FACTPICO benchmark, a framework aimed at evaluating how well medical summaries generated by LLMs like GPT-4, LLaMA-2, and Alpaca align with factual information using the PICO (Population, Intervention, Comparator, Outcome) structure. Their results showed a trade-off between factual accuracy and linguistic fluency: Alpaca was more accurate but less coherent, while LLaMA-2 and GPT-4 were more fluent but prone to errors, with hallucination rates of LLaMA-2 reaching up to 38\%. The PICO framework reflects a structured reasoning. Clinicians apply this structured reasoning when evaluating medical evidence. Thus, PICO-based evaluations capture high-level factual alignment between summaries and source material. Again, as PICO does not decompose text into fine-grained clinical claims, it struggles to detect instance-level inconsistencies like incorrect laboratory values, wrong dosages, omitted treatments, or temporal mismatches. That is why the PICO-based evaluation aligned better with clinician judgments, but struggled to detect inconsistencies at the instance level.

Hegselmann et al. \cite{hegselmann2024data} fine-tuned LLaMA-2 and GPT-4 using hallucination-free training data from MIMIC-IV discharge summaries. Their focus on a data-centric strategy, which includes annotated datasets with token-level hallucination labels and comprehensive evaluation processes, substantially lowered factual errors yet preserved clinical content. However, this approach remains primarily data-centric and does not directly address independent factual verification, leaving residual hallucination risks.

Garcia-Fernandez et al. \cite{garcia2025check} presented CHECK, a continuous learning approach that combines information-theoretic classifiers with specially designed clinical databases to address hallucinations in large models such as LLaMA3.3-70B-Instruct. When tested in clinical trial questions, CHECK improved the performance of GPT-4o on USMLE-like standards (achieving 92.1\%), decreased hallucination rates from 31\% to 0.3\% and had classifier AUCs of 0.95–0.96\%.

Regardless of its great reasoning abilities, Llama3.3-70B is computationally costly, difficult to fine-tune, and less practical for domain-specific research due to its massive parameter size \cite{garcia2025check, redhat2024llama3}. Llama3-8B, on the other hand, serves as a fair compromise between accuracy and efficiency, allowing faster adaptation to specialized medical datasets, such as MIMIC, without requiring unreasonably high processing resources \cite{dsssolutions2024benchmarking}.

Sawczyn et al. \cite{sawczyn2025factselfcheck} introduced a technique known as Fact-SelfCheck, which is a black-box, sampling-based framework that converts text into triple representations of knowledge graphs and checks for consistency between various model outputs. Compared to sentence-level methods that only improved factual accuracy by 8\%, this fact-level detection achieved a remarkable 35\% increase. This black-box nature and the reliance on model agreement rather than grounding in sources can limit its trustworthiness in clinical settings.

Hallucination detection without external resources has been further advanced by recent zero-knowledge detection frameworks like Finch-Zk \cite{goel2025zero} and Counterfactual Probing \cite{feng2025counterfactual}. Finch-Zk compares responses from multiple models for consistency to improve F1 detection by 6–39\%, while Counterfactual Probing dynamically employs slightly modified statements and analyzes model confidence shifts, which reduces hallucinations by about 24.5\% without the need for retraining. Identifying inconsistencies is the main focus of both methods, rather than fixing or establishing their foundation. Therefore, they can point out differences, but cannot truly guarantee that everything aligns with clinical facts.

\subsection{Automated Fact-Checking \& NLI-Based Verification}
Accuracy alone at the surface level does not guarantee the trustworthiness of generated content. Automated fact-checking and Natural Language Inference (NLI) based methods are important for verifying generated information with reliable sources \cite{vladika2023healthfc}, especially in medicine, where factual errors can endanger patient safety. Fact-checking goes far beyond just creating language models. The process entails breaking down the text into distinct claims and cross-referencing them with organized sources of knowledge or retrieved data \cite{kang2023evidencemap}.

Thorne and Vlachos \cite{thorne2018automated} provided a functional survey of automated fact-checking, highlighting challenges such as claim ambiguity, evidence attribution, and multi-step reasoning. Although fundamental, these methods were designed for open-domain tasks and lack direct applicability to clinical contexts requiring domain-specific reasoning.

Kazemi et al. \cite{quelle2024perils} explored the ReAct framework to evaluate GPT-3.5 and GPT-4 in fact-checking with and without external evidence. In particular, when contextual evidence was dynamically recovered, GPT-4 showed a significant improvement over GPT-3.5. However, this framework remains LLM-dependent for both reasoning and verification, making it vulnerable to evaluator-induced hallucinations and limiting reproducibility.

Graph-based fact-checking techniques are becoming more popular, such as MiniCheck \cite{tang2024minicheck}, and GraphCheck \cite{chen2025graphcheck}. MiniCheck decomposes documents and statements into individual claims, while GraphCheck enhances model inputs by incorporating extracted knowledge graphs, which helps with multihop logical reasoning. In both the general and medical fields, the approaches performed better than the baseline models. However, these methods rely on LLM evaluation and black-box reasoning, reducing transparency and reproducibility in clinical applications.

Similarly, Košprdić et al. \cite{kosprdic2024scientific} developed a DeBERTa-based NLI model trained on the HealthVer dataset. This model managed to verify health-related claims with a weighted F1 of 0.44 and accuracy of 0.50, while GPT-4 models outperformed others in entailment-based evaluations. Even with these advances, overall performance remains modest, and reliance on opaque model reasoning limits interpretability and reproducibility, potentially undermining clinician trust.

\subsection{Evaluation and Hallucination Mitigation Strategies}
Evaluating hallucinations and creating mitigation strategies are crucial to making LLMs reliable in healthcare. Unchecked hallucinations can lead to suggestions that are misleading or incorrect, jeopardizing patient safety \cite{thirunavukarasu2023large}. As clinical text generation differs from open-domain generation, it requires eloquent language, rigorous factual accuracy, and logical coherence \cite{ji2023survey}.

Surveys such as those by Tonmoy et al. \cite{tonmoy2024comprehensive} highlight data-centric and model inference-based mitigation approaches, which include retrieval augmentation, self-reflection, RAG, rapid engineering, and uncertainty calibration. These methods often address surface-level factuality without ensuring multi-step logical consistency. Further reviews \cite{vykopal2024generative, huang2023survey} emphasize limitations in evaluation metrics, domain-specific datasets, and RAG implementations. These highlight the need for robust, interpretable verification tools in specialized sectors.

\begin{sidewaystable*}[]
\centering

\caption{Comparison of prior work on hallucination detection, factuality evaluation, and medical fact-checking, highlighting task setting, granularity, verification mechanisms, reliance on LLMs, and limitations relative to the proposed proposition-level.}
\scriptsize
\begin{tabular}{|p{1.8cm}|p{2.2cm}|p{3.2cm}|p{2.7cm}|p{2cm}|p{3.5cm}|p{4cm}|}
\hline
\textbf{Work (Refs)} &
\textbf{Primary setting / task} &
\textbf{Granularity} &
\textbf{Verification mechanism} &
\textbf{Use of LLMs in verification} &
\textbf{Key strengths} &
\textbf{Main limitations} \\
\hline
% Work (Refs)
Maynez et al. \cite{maynez2020faithfulness} & 
Abstractive summarisation faithfulness (news) & 
Summary–source level & 
Comparison between human-rated faithfulness and automatic metrics & 
No LLM-as-verifier (pre-LLM era)& 
Establish “faithfulness vs factuality” distinction; show limits of lexical overlap metrics & 
Non-clinical domain; no proposition-level checks; no EHR or structured evidence \\
\hline
Med-HALT (Pal et al. \cite{pal2023med}) & 
Medical domain hallucination testing for LLMs & 
Question–answer / prompt-level & 
Benchmark that probes hallucinations on medical knowledge& 
Uses LLMs as the evaluated models; detection is via benchmark design& 
Medical-focused benchmark; highlights failure modes of LLMs on clinical knowledge & 
No explicit EHR grounding; it does not provide a verification pipeline for real clinical documents or summaries \\
\hline

CHECK (Garcia-Fernandez et al. \cite{garcia2025check})  & 
Continuous hallucination detection and elimination for medical LLMs & 
Utterance / segment-level & 
Pipeline for ongoing hallucination detection and mitigation around LLM outputs& 
Yes (LLM-based components in the loop)& 
Medical focus; designs a dedicated hallucination detection framework & 
Relies on LLMs and neural components; less transparent and potentially resource-intensive compared to purely symbolic verification\\
\hline

RAG methods (Lewis et al. \cite{lewis2020retrievalaugmentedgeneration}, Shuster et al. \cite{shuster2021retrieval}) & 
Knowledge-intensive NLP / conversational agents & 
Sentence / passage-level & 
Retrieval-augmented generation: incorporate external documents at generation time& 
Yes (LLM generator + retriever) &  
Reduce hallucination by grounding outputs in retrieved evidence & 
Do not explicitly verify propositions post hoc; not tailored to EHR structure; still rely on LLM behaviour at inference time\\
\hline
FACTPICO (Joseph et al. \cite{joseph2024factpico}) & 
Factuality of medical evidence summarisation (plain-language)& 
PICO-level (Population, Intervention, Comparator, Outcome)& 
Align summaries with evidence via PICO fields and PICO-based metrics& 
Uses LLMs for summarisation; evaluation is PICO-based, not LLM-as-judge& 
Clinically meaningful PICO abstraction; better alignment with clinician judgments than generic metrics & 
Coarse-grained; many instance-level inconsistencies (dosage, time, comorbidities) invisible at PICO level; not designed for EHR-scale proposition checking\\
\hline
HealthFC (Vladika et al. \cite{vladika2023healthfc}) & 
Verification of health claims with evidence-based fact-checking& 
Claim-level& 
Retrieve evidence and classify claim (supported/refuted)& 
Neural models (NLI-style)& 
Explicit health-claim verification; uses evidence-based medicine & 
Focuses on public health claims rather than patient-specific EHR summaries; no structured EHR proposition model\\
\hline

MiniCheck (Tang et al. \cite{tang2024minicheck})& 
Efficient fact-checking of LLMs on grounding documents& 
Sentence / span-level& 
Lightweight document-aware checking of LLM outputs& 
Uses LLMs and neural encoders& 
Efficient, document-grounded verification; scalable across tasks & 
General-purpose; not tailored to clinical ontologies or EHR structure; relies on neural components\\
\hline

GraphCheck (Chen et al. \cite{chen2025graphcheck})& 
Long-term text fact-checking with extracted knowledge graphs& 
Graph / triple-level& 
Build a knowledge graph and perform graph-powered checks& 
Neural extraction and graph reasoning& 
Captures long-range dependencies and structured relations & 
Extraction and reasoning pipelines are complex, not specifically clinical; continued reliance on neural components\\
\hline
DOSSIER (Zhang et al. \cite{pmlr-v252-zhang24a}) & 
Privacy-preserving fact-checking for EHRs& 
Sentence / segment-level& 
Dual-layer neural verification + entailment + retrieval; privacy-preserving architecture& 
Yes (neural verifiers and entailment models)& 
Explicitly designed for EHR and privacy; strong protection of patient data & 
Resource-intensive, opaque, and harder to reproduce, the neural + cryptographic stack reduces transparency; still vulnerable to model drift\\
\hline
BrainLLaMA / GPT-based evaluators (Siino et al. \cite{siino2024brainllama}, Siino \& Tinnirello et al. \cite{siino2024gpt})& 
LLM-based hallucination detection in benchmarks (e.g., SemEval)& 
Sentence / error-type level& 
LLM-as-judge with prompt engineering / specialised prompting& 
Yes (LLM-as-evaluator)& 
Flexible; easy to adapt prompts; competitive in shared tasks & 
Non-deterministic, costly, and not easily auditable; privacy and reproducibility concerns in clinical deployment\\
\hline

FactSelfCheck, Cross-model consistency, Counterfactual probing \cite{sawczyn2025factselfcheck, goel2025zero, feng2025counterfactual} & General LLM hallucination detection and mitigation& 
Fact / sentence-level &
Black-box detection via fact-level probing, cross-model consistency, and counterfactual queries& 
Yes (multiple LLMs or NLI-style models)& 
Do not require ground-truth documents; model-agnostic detection&
Not EHR-specific; no explicit use of structured clinical evidence; limited interpretability for clinicians\\
\hline

This work (proposition-level, LLM-free fact-checker) & Clinical summarisation verification against EHR& 
Proposition-level (entity–attribute–value–time)&
Deterministic mapping to propositions + numerical, temporal, logical, and presence checks& 
No LLMs in verification (LLM only for generation)& 
EHR-grounded, LLM-free verification; interpretable rule-based checks; suitable for privacy-preserving, auditable deployment&
Rule coverage and robustness are still limited in rare/specialised domains; it depends on the quality of proposition extraction and ontologies\\
\hline
% -------- Add more rows below --------
\end{tabular}
\label{tab:related_work} 
\end{sidewaystable*}
% ________________________END_________________________________

\subsection{Domain-Specific Error Detection and Correction}

In clinical NLP, contextual error detection and correction are indispensable. The top performer in the MEDIQA-CORR 2024 shared challenge was achieved using ensembles of LLMs (GPT-3.5, GPT-4, Claude), NER tools, and knowledge graphs (MeSH) \cite{benabacha2024overview}. However, these techniques frequently rely on external retrieval and LLMs, which limit scalability and interpretability. Kim et al. (2025) highlighted that minor fabricated details in clinical prompts can trigger hallucinations and emphasized the need for systematic verifications at the proposition-level in medical LLMs \cite{kim2025medical}. 

Zhang et al. \cite{pmlr-v252-zhang24a} introduced DOSSIER, a privacy-preserving fact-checking framework for Electronic Health Records (EHRs) that uses dual-layered neural verification, entailment-based checks, and retrieval-augmented verification. However, DOSSIER relies heavily on large neural components and complex cryptographic protocols, making it resource-intensive, opaque, and potentially less reproducible, with risks of hallucinations and model drift. These limitations point to the need for lighter-weight and more transparent fact-checking pipelines, for example, by combining secure data-access mechanisms with simpler, modular verifiers whose behavior can be systematically audited. In our work, we follow this direction by replacing neural verifiers with discrete logical checks over structured EHR propositions, while remaining compatible with privacy-preserving deployment. More broadly, future frameworks could explore distillation of verification models, hybrid neural–symbolic architectures, or standardized evaluation protocols to improve reproducibility while preserving strong privacy guarantees in EHR-scale deployments. 

% Zhang et al. \cite{pmlr-v252-zhang24a} introduced DOSSIER, a privacy-preserving fact-checking framework for Electronic Health Records (EHRs) that uses dual-layered neural verification, entailment-based checks, and retrieval-augmented verification. But DOSSIER relies heavily on neural networks and complex encryption, making it resource-intensive, opaque, and potentially less reproducible, with risks of hallucinations and model drift.

% Researchers have been diving into LLM-based evaluator systems in recent times, such as BrainLlama, which achieved accuracies of 0.62 (model-agnostic) and 0.67 (model-aware) in SemEval-2024 \cite{siino2024brainllama}, while Mistral-7B reached 0.73 (English) and 0.76 (Swedish) in the ELOQUENT 2024 Hallucination Detection task \cite{siino2024gpt}. 

Complementary to these EHR-focused frameworks, researchers have also been diving into LLM-based evaluator systems in recent times, such as BrainLlama and SemEval. BrainLlama achieved accuracies of 0.62 (model-agnostic) and 0.67 (model-aware) in SemEval-2024 \cite{siino2024brainllama}, while Mistral-7B reached 0.73 (English) and 0.76 (Swedish) in the ELOQUENT 2024 Hallucination
Detection task \cite{siino2024gpt}. Despite adaptability, these systems are prompt-sensitive and inconsistent across domains.

Although frameworks like DOSSIER improve verification and BrainLlama or Mistral-7B enhance evaluation adaptability, their dependence on LLM reasoning can create some opacity. Our proposed LLM-free validation framework uses structured EHR data and logical verification to ensure transparency, reproducibility, and reliability.

%After all this time, most current methods still rely on LLM-based verification, which unfortunately reintroduces hallucinations. Structured graph-based methods like NLI provide transparency but often lack domain adaptability. Though CHECK and FactSelfCheck are promising domain-specific models, general medical summarization systems without LLMs remain largely unexplored.

%Our LLM-free verification pipeline merges similarity search and propositional logic with a generator fine-tuned on MIMIC-III. By removing additional LLM dependency and enforcing proposition-level factual consistency, we aim to provide a structured, transparent, and reliable framework for clinical generation and evaluation.

Table~\ref{tab:related_work} summarises the main characteristics of prior work on hallucination detection, factuality evaluation, and medical fact-checking in relation to our setting. Surveys \cite{rawte2023survey}\cite{ji2023survey}–\cite{huang2023survey} provide high-level taxonomies and highlight hallucination as a central challenge, while methods such as Med-HALT \cite{pal2023med}, FACTPICO \cite{joseph2024factpico}, and HealthFC  \cite{vladika2023healthfc} focus on medical benchmarks and claim verification without operating directly on EHR-derived structures. Frameworks such as CHECK \cite{garcia2025check}, DOSSIER \cite{pmlr-v252-zhang24a}, MiniCheck \cite{tang2024minicheck}, GraphCheck \cite{chen2025graphcheck}, and LLM-based evaluator systems \cite{siino2024brainllama}, \cite{siino2024gpt} demonstrate powerful neural or LLM-centric verification pipelines, but often remain opaque, computationally intensive, and less suited to privacy-preserving deployment at EHR scale. In contrast, our approach is designed to complement competitive LLM-based summarisation with an LLM-free, proposition-level verification module grounded in structured EHR propositions, aiming to maximise transparency, reproducibility, and clinical interpretability.

\begin{figure}[]
\centering
\includegraphics[scale=0.07]{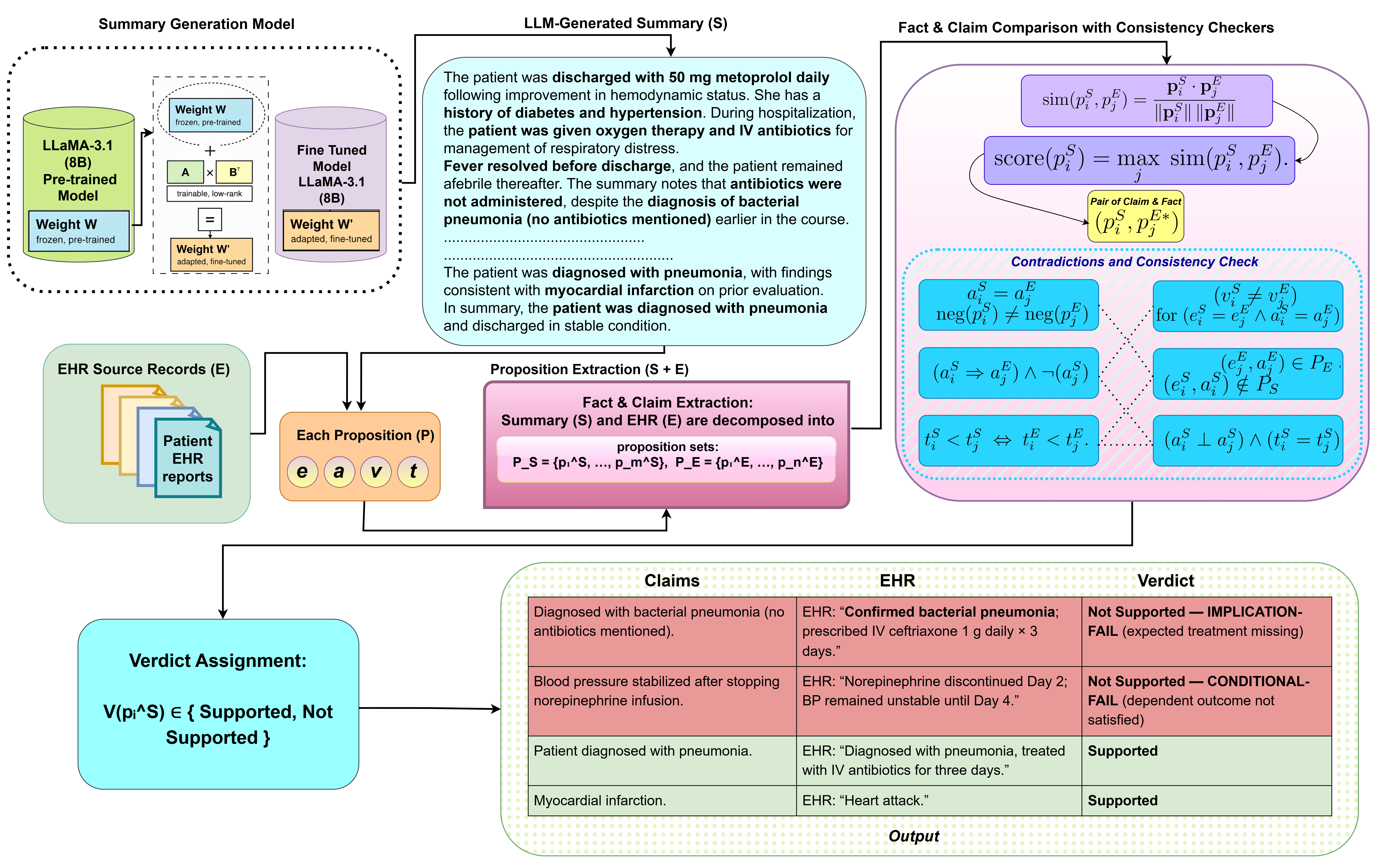}
\caption{The workflow illustrates how LoRA fine-tuning is applied to a large language model (LLM) to generate medical summaries from patient EHRs. The generated summaries and EHR source records are decomposed into structured propositions, which are then compared using consistency checkers to evaluate factual alignment. Contradictions and unsupported claims are identified through the logical consistency rules, which later result in a verdict assignment for each proposition (Supported or Not Supported).}
\label{fig: full-diagram}
\vspace{-15pt}
\end{figure}

\section{Methodology}

The complete architecture of our proposed work is illustrated in Figure \ref{fig: full-diagram}. It is designed to be modular and can be used with any LLM to assess the accuracy of the summaries. Our proposed work consists of a medical domain-specialized summarization model that is capable of producing factually correct summaries compared to other general models in this domain. It also includes a fact-checking module designed with discrete logic and various consistency checks, performed without the use of an LLM. These two primary modules are integrated to form a system that functions reliably. Each of the modules is independent; however, the proposed method is considered an effective system if used together. The verification module works as a layer to validate the summaries against the ground truths, which are the respective EHRs. This minimizes hallucinations and makes LLM-generated summaries reliable for medical decisions.

The evaluation framework assesses clinical accuracy at the level of discrete atomic units of the summaries, rather than complete sentences or entire documents. These units are called ``propositions." They include specific statements such as ``patient diagnosed with pneumonia," ``prescribed 20 mg lisinopril daily," or ``hemoglobin measured at 8.2 g/dL." In this way, our method can compare them systematically one by one. Both generated summaries and source EHRs are converted into propositions. This also ensures 
that the fact-checking module operates independently and can verify any summary generated by an LLM. However, our trained LLM performs better in generating summaries that are factually correct. The generator starts with a summary of clinical notes. After that, both the EHR and the text that was written are broken down into structured propositions. We then use different consistency metrics, such as lexical-semantic similarity, numerical cohesiveness, and temporal and logical consistency, to evaluate these propositions. Each proposition gets a verdict that states whether it is supported or not.

This evaluation is performed entirely without the help of LLMs, guaranteeing robustness and adaptability by employing medical synonyms as necessary and maintaining consistency in lab values, doses, and diagnoses, both logically and numerically.

\subsection{Medical Summary Generation}
We carried out a few initial experiments to develop and test a domain-specific summarization model that could turn raw EHR narratives into clinically coherent discharge summaries. We prepared the dataset, pre-processed it, and changed the LLaMA-3.1 (8B parameters) model to use parameter-efficient fine-tuning.

\subsubsection{Parameter-Efficient Adaptation Using Low-Rank Adaptation (LoRA)}

Fine-tuning all parameters of billion-scale large language models (LLMs) such as LLaMA-3.1 (8B parameters) is computationally expensive and memory-intensive, particularly for domain-specific applications. To mitigate these constraints, we adopt LoRA, a parameter-efficient fine-tuning approach that enables domain adaptation by introducing a limited number of trainable parameters while keeping the original pre-trained weights frozen. LoRA achieves this by incorporating low-rank decomposition within selected linear layers, effectively reducing training cost without degrading model performance. We specifically adopt LoRA over other parameter-efficient methods, such as QLoRA, because our experimental setup did not require 4-bit quantization to resolve memory constraints, and LoRA avoids potential accuracy losses introduced by low-bit quantization. LoRA provides stable, full-precision training dynamics, which are beneficial in domain-specialized settings. Moreover, LoRA’s simplicity makes it well-suited for our fine-tuning pipeline.

\paragraph{Theoretical Formulation of LoRA:}
Consider a linear transformation in a pre-trained neural model defined by a weight matrix $W \in \mathbb{R}^{d \times k}$. In conventional fine-tuning, all entries of $W$ are updated during training. LoRA, however, introduces a trainable low-rank update $\Delta W$, while preserving the original weights as static, defined in Equation (\ref{eqn:1}):
\begin{equation}
\label{eqn:1}
    W' = W + \Delta W, \qquad \Delta W = A B^{\top}
\end{equation}
Here, $W$ denotes the frozen pre-trained weight matrix. $A$ and $B$ are low-rank matrices that are learned during fine-tuning, where $A \in \mathbb{R}^{d \times r}$ and $B \in \mathbb{R}^{k \times r}$ with $r \ll \min(d, k)$. The hyperparameter $r$ represents the rank of the decomposition and governs the additional parameter budget. 

To control the magnitude of the learned adaptation, a scaling coefficient $\alpha$ is introduced in Equation (\ref{eqn:2}):
\begin{equation}
    W' = W + \frac{\alpha}{r} A B^{\top}
    \label{eqn:2}
\end{equation}

This formulation drastically reduces the number of trainable parameters from $\mathcal{O}(dk)$ to $\mathcal{O}(r(d + k))$, allowing efficient adaptation even for very large models. In practice, $r$ is typically selected between four and 16, and $\alpha$ between eight and 32, balancing adaptation flexibility and training stability.

\paragraph{LoRA Optimization Procedure:}

The LoRA-based fine-tuning process can be described as a sequence of efficient optimization steps. Initially, all pre-trained model parameters $W$ are frozen to preserve general linguistic and semantic knowledge obtained from large-scale pre-training. LoRA then introduces trainable low-rank matrices $A$ and $B$ into targeted layers. Most commonly, the query ($W_q$) and value ($W_v$) projection matrices in the multi-head attention mechanism. This selective adaptation ensures that only the components most responsible for contextual reasoning are fine-tuned for domain specialization.

During each forward pass, the adapted weight $W'$ is applied to the input $x$.
Substituting the LoRA parameterization from Equation ~(\ref{eqn:2}) utilized in Equation (\ref{eqn:3}):
\begin{equation}
    y = W x + \frac{\alpha}{r} A (B^{\top} x)
    \label{eqn:3}
\end{equation}

where $x$ represents the input of the layer. This formulation enables LoRA to learn task-specific residual updates while maintaining the integrity of the pre-trained representations. Gradients are propagated only through the matrices $A$ and $B$, leaving the frozen weights $W$ untouched. This selective gradient propagation significantly reduces computational overhead and memory consumption compared to full fine-tuning.

The optimization process is carried out using the AdamW optimizer with a learning rate ranging between $1 \times 10^{-4}$ and $5 \times 10^{-5}$, depending on the complexity of the dataset and the stability of the convergence. To further enhance computational efficiency, mixed-precision training (fp16/bf16) is used, and a batch size between 4 and 16 is typically sufficient to ensure stable convergence without overfitting. Upon completion of the fine-tuning, the learned low-rank parameters are merged with the base model using the same transformation defined in Equation~(\ref{eqn:2}), allowing deployment without additional adapter components. This merging step ensures that the inference speed and memory footprint remain equivalent to the original model, preserving real-time applicability.

\paragraph{Integration in Clinical Summarization:}

In this study, LoRA was applied to fine-tune the \textbf{LLaMA-3.1 (8B)} model using more than 40,000 patient records from the \textbf{MIMIC-III} critical care database to generate clinically coherent discharge summaries. Adapters were selectively integrated into the attention sub-layers, resulting in fewer than 1\% of the model’s parameters being updated during fine-tuning.

This efficient adaptation allowed stable convergence within a few epochs on limited hardware resources while preserving the fluency of the model and contextual accuracy. The resulting model demonstrated better factual grounding, reduced hallucination rates, and improved alignment with the structured content of Electronic Health Records (EHRs). These findings confirm that LoRA provides a scalable and computationally feasible pathway to fine-tune large medical language models in resource-constrained research environments.

\subsubsection{Optimization Setup}
The fine-tuning objective is to minimize the negative log-likelihood (NLL) of generating target tokens given the input sequence, as shown in Equation (\ref{eq:loss}):
\begin{equation}
\mathcal{L}(\theta) 
= - \sum_{i=1}^{N} \log P_{\theta}(y_i \mid x_{i}, \ldots, x_{1}),
\label{eq:loss}
\end{equation}
where $\theta$ represents the adjusted parameters, $x_i$ the input tokens and $y_i$ the corresponding target tokens. We used the AdamW optimizer with a cosine scheduler for the learning rate. Gradient accumulation was applied to simulate larger batch sizes.

\paragraph{LoRA Pseudocode:}
\begin{algorithm}[]
\caption{LoRA Fine-Tuning Procedure.}
\label{alg:lora}
\begin{algorithmic}[1]
\Require Pre-trained model $\mathcal{M}$ with parameters $\{W\}$, dataset $\mathcal{D}$, target layers $\mathcal{T}$ (e.g., $W_q, W_v$), rank $r$, scaling factor $\alpha$, optimizer hyperparameters (lr, epochs, batch\_size)
\Ensure Fine-tuned low-rank adapters $\{A, B\}$ and merged weights $W'$

\State \textbf{Initialization:}
\ForAll{modules $W \in \mathcal{M}$}
    \State Freeze base weight $W$ \Comment{preserve pretrained parameters}
\EndFor
\ForAll{$W \in \mathcal{T}$}
    \State Initialize $A \in \mathbb{R}^{d\times r}$, $B \in \mathbb{R}^{k\times r}$ (e.g., $\mathcal{N}(0,\sigma^2)$)
\EndFor
\State Initialize optimizer (AdamW) over adapter parameters $\{A,B\}$

% \Statex
\State \textbf{\textit{Training Phase:}}
\For{$e = 1$ \textbf{to} epochs}
    \For{each mini-batch $B_x \subset \mathcal{D}$}
        \State $L \gets 0$
        \For{each sample $x \in B_x$}
            \ForAll{target layer $W \in \mathcal{T}$}
                \State $z \gets B^{\top}x$
                \State $\Delta y \gets \frac{\alpha}{r} A z$
                \State $y \gets W x + \Delta y$
            \EndFor
            \State Accumulate loss $\ell(x)$ into $L$
        \EndFor
        \State $L \gets L / |B_x|$
        \State Backpropagate gradients (only for $\{A,B\}$)
        \State Update $\{A,B\}$ via AdamW optimizer
    \EndFor
\EndFor

% \Statex
\State \textbf{\textit{Merging Phase (optional):}}
\ForAll{$W \in \mathcal{T}$}
    \State $W' \gets W + \frac{\alpha}{r} A B^{\top}$ \Comment{merge adapters into base weights}
\EndFor
\State \Return merged weights $\{W'\}$ (or retain $\{A,B\}$ for modular adapters)
\end{algorithmic}
\end{algorithm}

For clarity and reproducibility, Algorithm~\ref{alg:lora} presents a concise pseudocode description of the LoRA fine-tuning workflow, adapted from Hu et al. \cite{hu2022lora}. The pseudocode follows the mathematical formulation introduced in Equation~(\ref{eqn:2}) and the optimization procedure described earlier.

\subsection{Fact Checking Module}  
The fact-checking module consists of various methods to ensure consistency checks. All the steps of verification operate on the structured propositions that were extracted both from the LLM-generated summary and the ground truths from the corresponding EHRs. These checks are constructed in such a way that it is able to detect any discrepancies or factual incorrectness in the summaries, such as numerical correctness, temporal alignment, and other rationales compared to the source data. The module functions by following a series of deterministic stages, which begin with extracting appropriate propositions, and then by logical rule-based evaluation. Finally, a verdict is generated after all the checks have been applied to each of the propositions. Each stage is independent of any large language reasoning model, which makes the process transparent. 

\subsubsection{Fact and Claim Extraction}  
 Each proposition is formally characterized as a tuple as in Equation (\ref{eqn:fact-extract}). In this representation, e denotes a clinical entity such as a medication or laboratory test, a denotes its attribute (for example, a diagnosis, prescription, or measurement), v captures the corresponding value, and t marks the time reference. This structure helps to ensure that each factual statement extracted from both the summary and the EHR is complete and clearly defined. Without these components, propositions can become incomplete, underspecified, or misleading in a clinical context. For example, just noting the entity “metoprolol” without the attribute “dosage” or the value “50 mg daily” does not represent the full clinical meaning. Similarly, omitting the temporal marker (“on discharge”, “on day 2”) can cause problems for time-sensitive conditions or treatments.

\begin{equation}
p = (e, a, v, t),
\label{eqn:fact-extract}
\end{equation}  

In practice, we extract $(e, a, v, t)$ from both the generated summary $S$ and the corresponding EHR document $E$ using a deterministic rule-based pipeline. First, each document is segmented into sentences. A clinical named-entity recognizer then identifies spans corresponding to diagnoses, procedures, medications, and laboratory tests. These spans are normalized to standard biomedical concepts (for example, SNOMED-CT, RxNorm, LOINC) using BioPortal \cite{noy2009bioportal}, which provides the entity field $e$ and ensures consistent concept normalization across documents.

For each sentence, we infer an attribute $a$ by applying dependency-based patterns and lexical cues such as “diagnosed with”, “treated with”, “started on”, “underwent”, or “lab value”. The value $v$ is extracted from the local context of each entity–attribute pair, combining numeric expressions (for example, “50 mg”, “38.5 °C”), qualitative labels (“positive”, “elevated”), and frequency or duration phrases (“twice daily”, “for three days”). Temporal markers $t$ are derived from explicit time expressions (“on day 2”, “before discharge”, calendar dates) and from document structure (admission and discharge times), which are normalized to a patient-specific timeline.

% where each $p_i$ represents a minimal factual statement extracted from either the summary or the EHR. Equation (7) assigns a verdict to each summary proposition based on its correspondence with EHR-derived propositions:  
% \begin{equation}
% V(p^S_i) \in \{ \text{Supported}, \text{Not Supported} \}.
% \end{equation}  
If a sentence mentions multiple distinct (entity, attribute, value, time) combinations, it is split into several atomic propositions so that each proposition $p$ corresponds to exactly one factual claim. Formally, the summary $S$ and EHR $E$ are decomposed into sets of atomic propositions, as shown in Equation~(\ref{eqn:prop-sets}):
\begin{equation}
P_S = \{ p^S_1, \ldots, p^S_m \}, \quad
P_E = \{ p^E_1, \ldots, p^E_n \},
\label{eqn:prop-sets}
\end{equation}  

For reliability, we did not use the extracted propositions directly; instead, we involved clinicians to evaluate the generated proposition sets. The clinicians verified whether the extracted (e, a,v,t) tuples correctly captured the clinical context of the original text, especially in more sensitive cases. This additional human-in-the-loop validation step helped us reduce the noise and ensure the accuracy of clinical contexts. Further evaluation details have been discussed in section \ref{sec:eval_details} as well as Table \ref{eval2}.

This decomposition makes the mapping from text to propositions explicit and reproducible, and the pipeline remains fully non-LLM-based, which also means any clinical NER and temporal tagger can be substituted in this step without changing the downstream verification logic.
\subsubsection{Fact Comparison}
Each summary proposition is compared with candidate propositions from the EHR, where the primary similarity is computed using cosine similarity. Each proposition $p^S_i$ searches for its designated factual counterpart in the EHR set $P_E$ by computing the cosine similarity between the embedding representations of $p^S_i$ and every $p^E_j$. The text of each proposition is first converted into dense vector embeddings using a fixed domain-specific biomedical encoder. In our experiments, we use BioClinicalBERT and Sentence-BERT because both are designed for sentence-level semantic similarity. BioClinicalBERT is pre-trained on MIMIC-III clinical notes and related biomedical corpora, which closely match the language and style of our EHR data, while Sentence-BERT fine-tunes transformer encoders with a similarity-oriented objective so that cosine distance correlates well with semantic relatedness. Compared to general-purpose encoders, these models better capture clinical synonymy and paraphrases (for example, ``myocardial infarction'' vs. ``heart attack'') and provide stable similarity scores without additional task-specific fine-tuning.

Although this step relies on pretrained encoders, no generative LLMs or probabilistic inference were used. Once the embeddings are computed, the selection of the best-matching EHR proposition for each summary proposition is purely deterministic. No further use of generative models is made in the verification module, which keeps the fact-checking process transparent and reproducible.

The cosine similarity between the summary and the embeddings of the EHR proposition is formally computed using its respective vector representations, as shown in Equation (\ref{eqn:cosine-sim}):
\begin{equation}
\text{sim}(p^S_i, p^E_j) = \frac{\mathbf{p}^S_i \cdot \mathbf{p}^E_j}{\| \mathbf{p}^S_i \| \, \| \mathbf{p}^E_j \|},
\label{eqn:cosine-sim}
\end{equation}  
where $\mathbf{p}^S_i$ and $\mathbf{p}^E_j$ denote the vector embeddings of the summary and EHR propositions, respectively, $\cdot$ represents the dot product, and $\| \cdot \|$ denotes the Euclidean norm.  The resulting similarity score has a range between $[0,1]$, and values closer to 1 indicate a higher semantic alignment.  

To compare $p^S_1$ with $p^E_3$, their vectorized representations are computed, and the EHR proposition with the highest cosine similarity is selected. This process aligns $p^S_1$ with the most semantically related factual statement in the EHR. The high similarity score effectively captures these medical synonyms. Once each pair of propositions $(p^S_i, p^{E*}_j)$ is identified by obtaining the maximum similarity, the pairs matched are then sent to the logical verification module for the next steps.
Formally, for each summary proposition, the maximum similarity score is computed on all EHR propositions as shown in Equation (\ref{eqn:max-score}):
\begin{equation}
\text{score}(p^S_i) = \max_{j} \; \text{sim}(p^S_i, p^E_j).
\label{eqn:max-score}
\end{equation}
This scoring mechanism ensures that each $p^S_i$ is paired with the factual counterpart that is the most semantically consistent $p^{E*}_j$ of the EHR set. 

However, cosine similarity serves purely as a candidate selection step to identify the most semantically aligned EHR proposition. It does not contribute to determining factual correctness. The verification decision is made independently through deterministic rule-based checks without any reliance on similarity thresholds or percentage-based deviation measures. No similarity threshold is applied in this step. In practice, LLM-generated summary propositions and EHR-derived propositions are typically well-aligned, and very low cosine similarity scores are rarely found during the process. However, even when the highest similarity score is relatively low (e.g., 0.2–0.3), the argmax criterion ensures that the most similar EHR proposition is selected for alignment. This selection does not imply factual acceptance. For example, a summary proposition such as “patient treated with antibiotics” may be aligned with an EHR proposition such as “patient underwent imaging study” if no closer match exists, resulting in a low similarity score (e.g., 0.3). Such a pair represents weak semantic correspondence and fails subsequent logical verification checks (e.g., presence or implication consistency). Consequently, the proposition is labeled as Not Supported. Thus, low-similarity alignments are not treated as valid matches, but they are systematically rejected during the deterministic verification stage.

\subsubsection{Contradictions and Consistency Check}
After each summary proposition $p^S_i$ has been aligned with its most similar EHR proposition $p^{E*}_j$, the pair is passed to a logical verification module. This module performs several complementary checks that jointly assess whether the two propositions can be true at the same time. We consider \textbf{negation}, \textbf{implication}, \textbf{temporal ordering}, \textbf{mutual exclusivity}, and \textbf{numerical consistency} on matched pairs, plus a presence check on unmatched EHR propositions. All summary claims pass through the same sequence of checks against their EHR ground truths. All consistency checks are applied to each proposition without skipping any of them. Below, we briefly describe each check and give an illustrative example.

% We designed our verification module by combining several components that makes the use of discrete logic as a multi-layered consistency checking mechanism. This systematically validates the factual, logical, and temporal integrity between the generated summary and the source EHR. This design performs five complementary checks that include \textbf{negation}, \textbf{implication}, \textbf{temporal ordering}, \textbf{mutual exclusivity}, and \textbf{numerical consistency}. Each is formally defined and operationalized within the pipeline where all the claims from the summary go through the same set of steps and logical checks against their ground truths. Together, these checks make sure that every extracted proposition is verified before being accepted as supported. The workflow for these sequential checks is illustrated in Figure~\ref{fig:fact-check-table}.

\begin{figure*}[]
    \centering
    \includegraphics[scale=0.07]{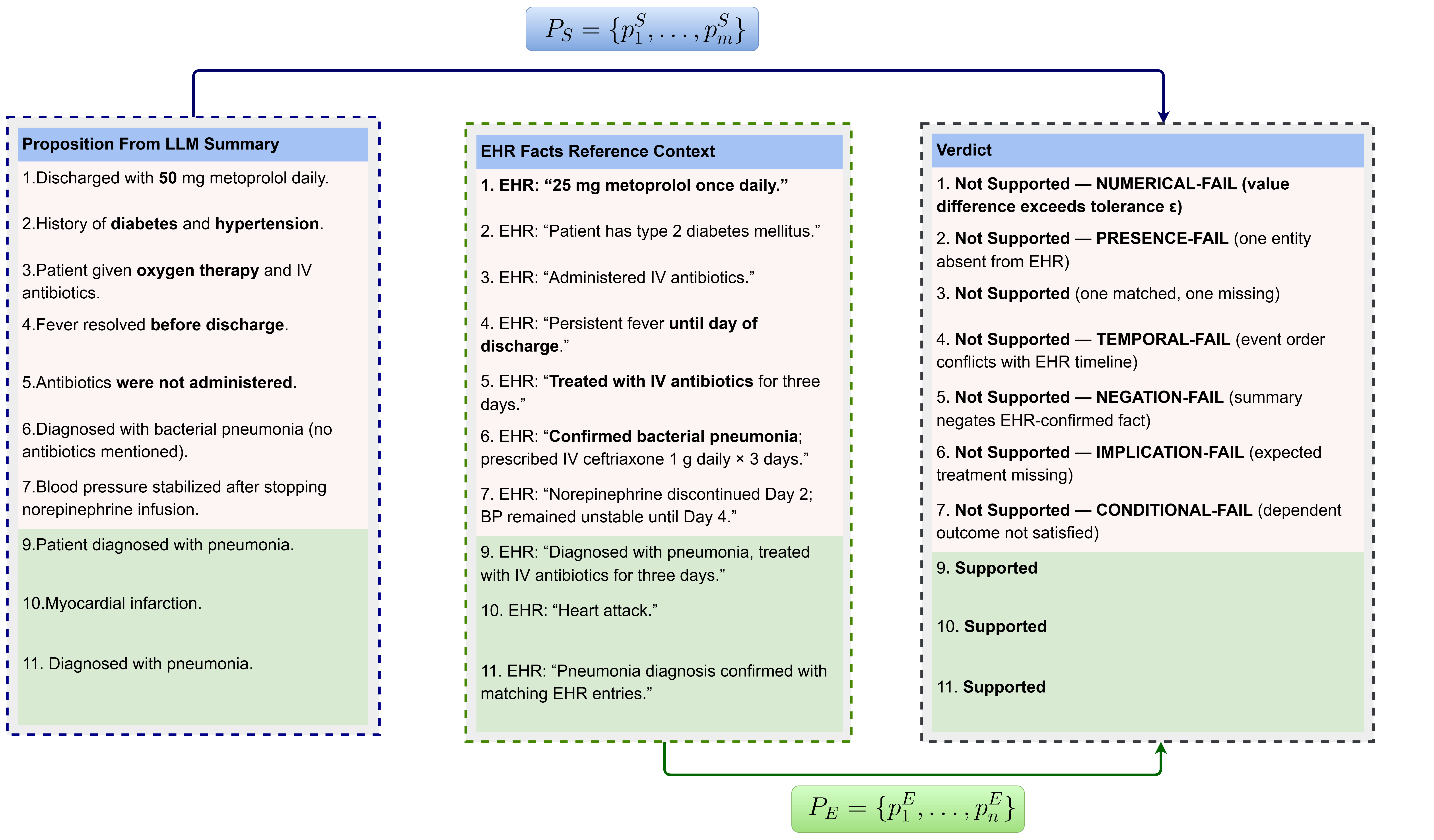}
    \caption{Illustrative examples of proposition-level factual verification outcomes are shown here. Each row shows a proposition based on a summary, its EHR reference statement, and the factual verdict that was given after running through the fact-checking module. The figure shows different kinds of consistency checks that the verification engine does. A few of the checks are shown here, such as numerical, presence, temporal, negation, implication, and mutual exclusivity checks. Propositions failing one or more checks are marked as \texttt{Not Supported}, and those fully aligned and validated against the EHR are labeled as \texttt{Supported}. These highlight the deterministic operation of the multi-layered fact-checking pipeline.}
\label{fig:fact-check-table}
    \vspace{-15pt}

\end{figure*}

\paragraph{Negation Check:}
 The negation check identifies direct contradictions between summary and EHR propositions. It flags cases where one claim asserts the presence of an event while the other asserts its absence, for the same entity and attribute. We infer the negation state of each proposition using lexical cues such as “no”, “not”, “denies”, “without”, or “ruled out”, and then require the negation polarity to match when e and a are the same. For any pair ($p^S_i$, $p^{E*}_j$), a negation failure is triggered when the attributes agree but the negation states differ (as formalized in Equation~(\ref{eqn:negation-fail})):

\begin{equation}
a^S_i = a^E_j \; \text{and} \; \text{neg}(p^S_i) \neq \text{neg}(p^E_j),
\label{eqn:negation-fail}
\end{equation}
Where $\text{neg}(\cdot)$ denotes the negation state of the proposition.\\
For example, if the summary states “antibiotics were not prescribed” while the EHR reports “treated with IV antibiotics for three days”, the pair is labeled NEGATION-FAIL Figure ~\ref{fig:fact-check-table}, (example 5). Similarly, “no evidence of pneumonia” in the summary and “treated for pneumonia” in the EHR also constitute a negation conflict.

\paragraph{Implication Check:}

The implication check verifies clinically dependent relationships defined by simple ontology-based rules. Certain diagnoses, procedures, or states imply that another event must have occurred (for example, “pneumonia $\Rightarrow$ antibiotics”, “mechanical ventilation $\Rightarrow$ intubation”). If the antecedent appears but its implied consequence is missing, we flag a logical inconsistency. Formally, if an attribute $a^S_i$ implies another attribute $a^E_j$ but no corresponding proposition for $a^E_j$ is present, an IMPLICATION-FAIL is raised (as seen in Equation~(\ref{eqn:implication-fail})):

\begin{equation}
(a^S_i \Rightarrow a^E_j) \wedge \neg(a^S_j) \Rightarrow \texttt{IMPLICATION-FAIL}.
\label{eqn:implication-fail}
\end{equation}
For instance, if a summary mentions “community-acquired pneumonia” but omits any antibiotic treatment while the EHR records “IV ceftriaxone for three days”, the implication rule “pneumonia $\Rightarrow$ antibiotics” is violated, as seen in Figure ~\ref{fig:fact-check-table}, example 6. This check captures omissions that are clinically implausible rather than directly contradictory.

\paragraph{Temporal Consistency Check:}
Temporal consistency ensures that the order and duration of medical events are chronologically aligned between the EHR and the summary. Using the normalized time markers $t^S_i$ and $t^S_j$ attached to each proposition, the system verifies that relative ordering is preserved: if the summary states that event i happened before event j, the same ordering must hold in the EHR (as expressed in Equation~(\ref{eqn:temporal-fail})):
\begin{equation}
t^S_i < t^S_j \; \Leftrightarrow \; t^E_i < t^E_j.
\label{eqn:temporal-fail}
\end{equation}
For example, if the summary says “fever resolved before discharge”, but the EHR shows “fever persisted until discharge”, a TEMPORAL-FAIL is raised (Figure ~\ref{fig:fact-check-table}, example 4). Likewise, a summary that claims “extubated before transfer to the ward” contradicts an EHR in which extubation occurred after the transfer. This check prevents subtle misrepresentations of clinical timelines.

\paragraph{Mutual Exclusivity Check:}

Some clinical states cannot logically co-occur at the same time. The mutual exclusivity check flags incompatible propositions that are assigned identical or overlapping time markers. If two attributes $a^S_i$ and $a^S_j$ are known to be mutually exclusive but share the same time $t^S$ (Equation~(\ref{eqn:exclusivity-fail})), an EXCLUSIVITY-FAIL is raised.
% Certain medical states or events cannot logically co-occur. The mutual exclusivity check flags incompatible propositions that appear simultaneously, as shown in Equation~(\ref{eqn:exclusivity-fail}):
\begin{equation}
(a^S_i \perp a^S_j) \wedge (t^S_i = t^S_j) \Rightarrow \texttt{EXCLUSIVITY-FAIL},
\label{eqn:exclusivity-fail}
\end{equation}

For example, “intubated” and “spontaneously breathing on room air” recorded at the same time violate exclusivity, as do “NPO” (nothing by mouth) and “tolerating regular oral diet” for the same time interval. This check encodes simple but robust clinical constraints that are not captured by semantic similarity alone.

% where $(a^S_i \perp a^S_j)$ indicates mutually exclusive clinical states. For example, ``intubated'' and ``spontaneously breathed'' recorded at the same time violate exclusivity.

\paragraph{Numerical Consistency Check:}
Semantic similarity is not sufficient for quantitative statements, which must also match their numeric values and units. The numerical consistency check compares the value $v^S_i$ in the summary with $v^E_j$ in the EHR for the same entity and attribute (Equation~(\ref{eqn:numerical-fail})). A mismatch in value or unit triggers a  \texttt{NUMERICAL-FAIL}.

% Semantic similarity alone is insufficient because numerical propositions must also match their correct entities and attributes. For each pair $(p^S_i, p^{E*}_j)$, a mismatch is identified if the numerical values differ for the same clinical variable, as defined in Equation~(\ref{eqn:numerical-fail}):
\begin{equation}
(v^S_i \neq v^E_j) \; \text{for} \; (e^S_i = e^E_j \wedge a^S_i = a^E_j).
\label{eqn:numerical-fail}
\end{equation}
 For example, if the summary states “creatinine 1.2 mg/dL” while the EHR reports “creatinine 2.1 mg/dL” for the same time point, or if the summary reports “blood pressure 120/80” when the EHR shows “180/100”, the propositions are numerically inconsistent (Figure ~\ref{fig:fact-check-table}, example 1). This check ensures that detailed quantitative information is faithfully copied rather than loosely paraphrased.
 
% This ensures that each attribute-value pair is compared only to its corresponding variable (e.g., blood pressure to blood pressure, not to heart rate). Any deviation triggers a \texttt{NUMERICAL-FAIL}, as demonstrated in Figure ~\ref{fig:fact-check-table}, example (1)

Propositions that fail to pass one or more checks are marked as \texttt{Not Supported} and can be further passed to expert moderation. This layered verification process systematically finds, labels, and records factual and logical inconsistencies to ensure that the generated clinical summaries remain reliable and accurate for decision support.

\paragraph{Presence Check (Omission Detection):}
Finally, to detect clinically important omissions, we perform a presence check over all EHR propositions that do not have a high-similarity counterpart in the summary. If an EHR proposition ($(e^S_i, a^S_i)$) exists for a key diagnosis, treatment, or event but no corresponding pair ($e^S_i, a^S_i$) can be found in the summary, a \textsc{PRESENCE-FAIL} is raised (Equation~\ref{eqn:presence-fail}):
% Finally, to detect clinically important omissions, we perform a presence check over all EHR propositions that do not have a high-similarity counterpart in the summary. If an EHR proposition $(e^E_j, a^E_j)$ exists for a key diagnosis, treatment, or event but no corresponding pair $(e^S_i, a^S_i)$ can be found in the summary, a \textsc{PRESENCE-FAIL} is raised (Equation~\ref{eqn:presence-fail}).
\begin{equation}
(e^E_j, a^E_j) \in P_E \; \text{and} \; (e^S_i, a^S_i) \notin P_S \Rightarrow \texttt{PRESENCE-FAIL}.
\label{eqn:presence-fail}
\end{equation}
For example, if the EHR contains “IV antibiotics administered for three days” but the summary does not mention antibiotic therapy at all, the omission is flagged as a presence failure (Figure ~\ref{fig:fact-check-table}, example (2)). This check guards against errors of omission and ensures that essential clinical events are not silently dropped from the generated summary.

Propositions that fail one or more of these checks are labeled Not Supported, whereas those that pass all applicable checks are labeled Supported. The combined use of pairwise logical checks and the presence check enables the fact-checking module to capture both explicit contradictions and clinically important omissions at the proposition level.

\subsubsection{Verdict Generation}  
Each summary proposition receives a final verdict using the integrated outcomes of semantic, numerical, and logical validation. The verdict is assigned as shown in Equation (\ref{eqn:verdict})
\begin{equation}
V(p^S_i) =
\begin{cases}
\text{Supported}, & \text{if aligned and validated}, \\
\text{Not Supported}, & \text{otherwise}.
\end{cases}
\label{eqn:verdict}
\end{equation}  
Figure \ref{fig:fact-check-table} provides illustrations of this decision process, showing propositions that were verified as supported or flagged as not supported under various consistency checks, such as temporal, numerical, and implication violations. These examples demonstrate the operation of the fact-checking module.

All of these combine to form a fact-checking layer, which works as per the examples demonstrated. In this way, each proposition of the LLM-generated output can be verified effortlessly and can improve dependability in critical decision-making periods. 

\subsection{Verification System Integration}  
The verification pipeline functions as a post-evaluation layer that works independently of the language model. However, such layers can be implemented in other benchmark generative models. This layer ensures that generated summaries are validated against their EHR data. Each proposition is analyzed against its ground truth through the logical checks, and then a transparent verdict (\textit{Supported} or \textit{Not Supported}) is assigned. The demonstrated results in Figure \ref{fig:fact-check-table} highlight the range of factual and logical inconsistencies identified by the system in real-world clinical narratives.

\section{Experiments}

This section describes the experimental setup of our approach to the LLM-free clinical fact verification method and the summarization model. We outline the dataset pre-processing and model fine-tuning, as well as an evaluation methodology to measure factual, numerical, and logical coherence across generated summaries and their paired EHR records. The following subsections detail the dataset utilized, the pre-processing pipeline, the model training settings, the evaluation strategy, and how the verification system is applied in conjunction with automated and human validation. 
\subsection{Dataset}
\begin{table}[htbp]
\caption{Summary of MIMIC-III Clinical Database (Information collected from PhysioNet)}
\label{tab:mimiciii_summary}
\centering
\small
\renewcommand{\arraystretch}{1.2}
\scriptsize
\begin{tabular}{l l p{8cm}}
\hline
\textbf{Category} & \textbf{Attribute} & \textbf{Details} \\
\hline
\multicolumn{3}{c}{\textbf{General Info}} \\
\hline
General Info & Full Name & Medical Information Mart for Intensive Care III \\
 & Developed By & MIT Laboratory for Computational Physiology \\
 & Data Source Hospital & Beth Israel Deaconess Medical Center \\
 & Location & Boston, USA \\
 & Data Period & 2001--2012 \\
 & Data Type & De-identified health-related data associated with ICU patients \\
\hline
\multicolumn{3}{c}{\textbf{Patient Info}} \\
\hline
Patient Info & Age Group & Adult and neonatal patients \\
 & Demographics & Age, gender, ethnicity \\
\hline
\multicolumn{3}{c}{\textbf{Clinical Data}} \\
\hline
Clinical Data & Vital Signs & Yes \\
 & Lab Tests & Yes \\
 & Medications & Yes \\
 & Procedures & Yes \\
 & Clinical Notes & Yes \\
 & Time-Series Data & Yes \\
 & Imaging & Not included (radiology reports available as text) \\
\hline
\multicolumn{3}{c}{\textbf{Diseases \& Classes}} \\
\hline
Diseases \& Classes & Coding System & ICD-9 \\
\hline
\multicolumn{3}{c}{\textbf{Hospital / ICU}} \\
\hline
Hospital / ICU & Hospital Type & Single tertiary academic medical center \\
 & ICU Types & Multiple ICUs (e.g., medical, surgical, cardiac, neonatal) \\
 & Number of Hospitals & 1 \\
\hline
\multicolumn{3}{c}{\textbf{Database Structure}} \\
\hline
Database Structure & Core Tables & PATIENTS, ADMISSIONS, ICUSTAYS \\
 & Event Tables & CHARTEVENTS, LABEVENTS \\
 & Medication Table & PRESCRIPTIONS \\
 & Notes Table & NOTEEVENTS \\
 & Procedures Table & PROCEDURES\_ICD \\
\hline
\end{tabular}
\end{table}

Our experiments were conducted on the MIMIC-III database, a publicly available de-identified electronic health records (EHRs) dataset comprising more than 40,000 patients and widely used as a benchmark for clinical prediction tasks \cite{johnson_mimiciii_2016}. From this dataset, 26,104 discharge summaries were sampled to fine-tune, validate, and test the summarization model. These summaries contain detailed clinical narratives, making them well-suited for summarization tasks.

% \textcolor{blue}{The dataset includes both structured and unstructured clinical data collected from intensive care units, such as discharge summaries, physician and nursing notes, laboratory measurements, medication records, procedure information, and time-stamped chart events. It provides detailed patient-level information, including diagnoses, treatments, vital signs, and longitudinal clinical observations recorded throughout hospital stays, covering a wide variety of disease conditions encountered in critical care, including both chronic and acute diseases. Clinical conditions in MIMIC-III are encoded using the International Classification of Diseases, Ninth Revision (ICD-9), a standardized hierarchical coding system that groups diagnoses into broader categories such as respiratory, cardiovascular, infectious, metabolic, and neurological disorders. These groups reflect the diversity of clinical conditions in the dataset and should not be interpreted as the classification labels used in our fact-checking task.}

The dataset includes both structured and unstructured clinical data collected from intensive care units, such as discharge summaries, physician and nursing notes, laboratory measurements, medication records, procedure information, and time-stamped chart events. It provides detailed patient-level information, including diagnoses, treatments, vital signs, and longitudinal clinical observations recorded throughout hospital stays, covering a wide variety of clinical conditions encountered in critical care, including acute conditions and chronic comorbidities. Clinical conditions in MIMIC-III are encoded using the International Classification of Diseases, Ninth Revision (ICD-9), which is a standardized coding system that groups diagnoses into broader categories such as respiratory, cardiovascular, infectious, metabolic, and neurological disorders and many others. These groups reflect the diversity of clinical conditions in the dataset and should not be interpreted as the classification labels used in our fact-checking task. A summary of the dataset characteristics, including its scale, structure, and key components, is provided in Table \ref{tab:mimiciii_summary}.

For fact-checking evaluation, 104 patient discharge records were selected, from which a total of 3,786 propositions were extracted and validated. Although MIMIC-III contains over 40,000 encounters, this subset was used due to the requirement of detailed, proposition-level annotation by clinicians, which is highly time-intensive. The selected records were stratified to cover diverse diagnoses, care units, and document lengths, ensuring that the evaluation remains both representative and feasible.  This curated subset ensures that a wide range of disease categories are covered, which helps with representative and generalizable factuality assessment without adding noise to the annotations.

\subsection{Preprocessing}
We carried out preprocessing to ensure that the data were suitable enough and ready for the determined fine-tuning task. The data set was filtered to keep only the discharged notes that contained detailed narratives of patient records. Initially, we retrieved discharge summaries from the MIMIC-III dataset and paired them with their corresponding clinical notes. Records that were too brief or excessively lengthy were eliminated in the preprocessing of the dataset to keep everything consistent with the model’s input requirements.

In addition, a slight moralization was performed. This involved making the text all lowercase, eliminating any non-essential symbols, and standardizing common clinical abbreviations as necessary. We ensured that all the important clinical terms, numerical values, and units remained unchanged since they are essential for accurate fact-checking down the line. We made sure to eliminate any duplicate records and filtered out incomplete entries that lacked important clinical information, like diagnosis or treatment specifics. After all that careful filtering, we were left with 26,104 high-quality discharge summaries. Then, the dataset was partitioned into three subsets: training (20,883 samples), validation (2,610 samples), and test (2,611 samples). The division was done based on an 80:10:10 ratio, with a random seed value of 42.

\begin{table}[]
\centering
\scriptsize
\caption{Training Configurations.}
\label{tab:training_config}
\begin{tabular}{p{0.28\linewidth} p{0.62\linewidth}}
\hline
\textbf{Component} & \textbf{Configuration} \\
\hline
Optimizer & AdamW, Learning Rate = $1 \times 10^{-4}$, cosine schedule, weight decay = 0.01 \\
Batching & Effective batch size = 8 (1 per device $\times$ 8 grad accumulation) \\
Epochs / Steps & 1,200 steps ($\approx$ 2 epochs) \\
Precision & \texttt{bfloat16} mixed precision with gradient checkpointing \\
Loss Function & Token-level cross-entropy (masked prompts) \\
Trainable Params & $\sim$84M / 8.1B total (1.03\%) \\
\hline
\end{tabular}
\end{table}

\subsection{Training Setup}
During fine-tuning, AdamW was used as the optimizer with a learning rate of $1 \times 10^{-4}$, a cosine decay schedule, and weight-decay of 0.01. The batch size used in this study was 8, with a sequence length of 2,048 tokens. Mixed precision (bfloat16) and gradient checkpoint were used for better memory efficiency. The use of token-level masked cross-entropy loss encourages stable convergence in sequences of different lengths.

Table~\ref{tab:training_config} presents a complete list of hyperparameters and optimization details and reveals that the top configuration focuses on parameter efficiency through partial fine-tuning, thus it only has about 84 million trainable parameters (1.03\% of the total of 8.1 billion). This setup ensures reproducibility and easy large-scale model fine-tuning on a single A100 GPU with performance consistency.

\subsection{Training Pipeline}
\begin{figure}[]
    \centering
    \includegraphics[scale=0.075]{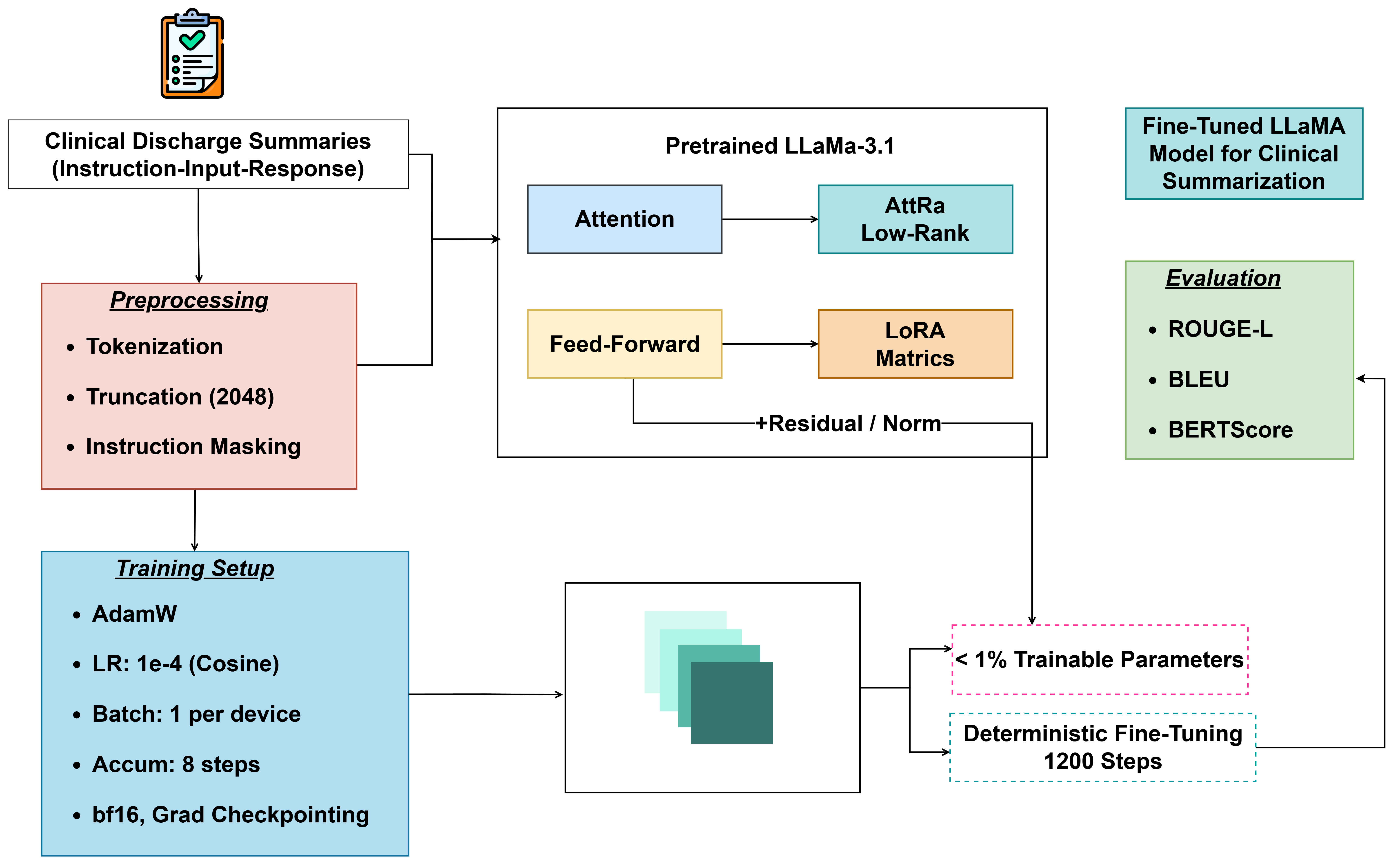}
    \caption{End-to-end fine-tuning pipeline illustrating preprocessing, tokenization, LoRA-based adaptation, and evaluation integration.}
    \label{fig:pipeline}
    \vspace{-10pt}
    
\end{figure}

Figure~\ref{fig:pipeline} shows the complete workflow, including preprocessing, tokenization, and alignment of the raw text into instruction-response pairs. Inputs were segmented for retrieval-augmented learning, and LoRA modules were inserted for adaptation. The system optimized the negative log-likelihood loss with validation monitoring and was tested on held-out datasets. This pipeline integrates data preparation, representation learning, adaptation, optimization, and evaluation into a single unified process.

LoRA adds only $\mathcal{O}(r(d+k))$ trainable parameters, whereas full fine-tuning adds $\mathcal{O}(dk)$. In our setup, $r=8$ reduced the trainable parameters by more than 99\%, enabling efficient training on a single A100 GPU.

\subsection{Evaluation Method}
We evaluated the correctness of each proposition of our fact-checking module using precision, recall, F1-score, and confusion matrix analysis. These metrics allow us to understand how well the generated summaries align with their respective EHR data. The MIMIC-III dataset provides both discharge summaries and structured EHRs, which allows us to match each summary with its actual record, which also ensures a solid factual assessment.

\subsection{Integration of the Verification System}
All summaries generated from the LLMs went through the verification pipeline's steps; each extracted proposition was logically compared using the proposed fact-checking module with its corresponding EHR proposition and labeled as either ”Supported” or ”Not Supported.” Then a clinician reviewed and checked these findings to ensure that there were no inaccuracies in the results. Since summary, EHR, and logical consistency are the main constants throughout the pipeline, this integrated approach keeps the process straightforward and reliable.

\subsection{Clinical Expert Evaluation}
\begin{table}[ht!]
\centering
\caption{Evaluation rubric used for experts' assessment of generated summaries and propositions}
\label{eval}
\scriptsize
\begin{tabular}{p{3.5cm} p{8.5cm} p{3cm}}
\toprule
\textbf{Dimension} & \textbf{Definition} & \textbf{Scale} \\
\midrule
Factual Accuracy & Measures whether clinical statements (or extracted propositions $p=(e,a,v,t)$) in the summary are fully supported by the source EHR without introducing incorrect or unsupported information. & 1 (Incorrect) -- 5 (Fully supported) \\
\addlinespace
Completeness & Assesses the extent to which the summary preserves clinically relevant information from the original record, including key diagnoses, treatments, findings, and their corresponding propositions. & 1 (Highly incomplete) -- 5 (Fully complete) \\
\addlinespace
\bottomrule
\end{tabular}
\end{table}

\begin{table}[ht]
\centering
\caption{Summary of experts' evaluation results on 104 generated clinical summaries and their propositions}
\label{eval2}
\scriptsize
\begin{tabular}{p{6cm} p{6cm}}
\toprule
\textbf{Metric} & \textbf{Value} \\
\midrule
Number of clinicians & 2  \\
Number of evaluated summaries & 104 (100\%) \\
Clinically consistent summaries & 97/104 (93.27\%) \\
Disagreement cases & 4/104 (3.85\%) \\
Ambiguous cases & 3/104 (2.88\%) \\
\midrule
\textbf{Proposition Extraction Quality (on annotated subset)} & \\
Proposition correctness & 95.8\% \\
Coverage of gold propositions & 92.9\% \\
Exact match rate (p = (e,a,v,t)) & 91.5\% \\
Common error types & Temporal attribute omissions; entity boundary errors; incorrect attribute--value pairing; missed propositions \\
\midrule
\textbf{Factual Accuracy (Score Distribution)} & \\
Score 5 & 65/104 (62.50\%) \\
Score 4 & 32/104 (30.77\%) \\
Score 3 & 5/104 (4.81\%) \\
Score 2 & 1/104 (0.96\%) \\
Score 1 & 1/104 (0.96\%) \\
\addlinespace
\textbf{Completeness (Score Distribution)} & \\
Score 5 & 60/104 (57.69\%) \\
Score 4 & 37/104 (35.58\%) \\
Score 3 & 5/104 (4.81\%) \\
Score 2 & 1/104 (0.96\%) \\
Score 1 & 1/104 (0.96\%) \\
\bottomrule
\end{tabular}
\end{table}

The clinicians evaluated summaries generated from MIMIC-III EHRs through a pairwise comparison of the output summaries against their respective source records. Overall, there was maximum agreement between the summaries and their respective source records, with some room left for improvement, particularly in capturing rare disease conditions and temporal dependency for extended periods of time. All evaluations are done on anonymized data in compliance with the MIMIC-III usage policies.

Human evaluation was carried out for clinical reliability assessment on 104 generated summaries with two clinicians experienced in reviewing electronic health records (EHR). A selection of diverse cases with varying degrees of complexity regarding different medical conditions was used for the purpose of this evaluation. Evaluation involved comparing generated summaries against their respective source records based on the following clinically relevant criteria, which are factual accuracy and completeness. As defined in Table \ref{eval}, the first checks if the summary factually matches the original record, and the second one checks if any important details are missing. The two clinicians independently reviewed all samples. A five-point Likert scale \cite{batterton2017likert} was used for all dimensions, supported by detailed annotation guidelines to ensure consistency across evaluators. Both clinicians evaluated all summaries independently. From analysis of agreement between the clinicians, the evaluation process was found to be highly reliable since disagreements were minimal, and any disagreements were minimal and resolved through discussion to obtain a consensus judgment used for final analysis. In addition, the same evaluation rubric was applied to assess the correctness and completeness of extracted propositions $p=(e,a,v,t)$ with respect to the source EHRs on a manually annotated subset.

As shown in Table \ref{eval2}, out of the total number of summaries evaluated, 97 out of 104 summaries (93.27\%) were found to be clinically accurate with the corresponding patient records. The few exceptions where disagreement occurred mainly revolved around complex clinical cases involving rare illnesses and temporal dependences.  All the summaries were evaluated anonymously in accordance with MIMIC-III requirements.

\label{sec:eval_details}
% Overall, this comparison highlights a key trade-off: while LLM-based fact checkers may provide broader coverage, they remain vulnerable to evaluator-induced hallucination and nondeterministic outputs. Our LLM-free pipeline demonstrates that competitive factual accuracy can be achieved with greater transparency, reproducibility, and efficiency, which makes it a more reliable choice for clinical summarization.
\section{Results}

In this section, we present the results of our LLM-free fact-checking module, which has been applied to summaries generated by an LLM-based system for clinical summarization, evaluated on the MIMIC-III dataset.
The LLM-free approach applies specifically to the verification component, which operates independently of any LLM; LLMs are used solely for summary generation. Thus, the proposed framework is LLM-based for generation but strictly LLM-independent for verification, and the claim of “LLM-free” refers only to the fact-checking stage.
The experiments evaluate the summarization quality of the LoRA-fine-tuned LLaMA-3.1-8B generator and the factual accuracy of the independent verification module across 104 discharge summaries, comprising 3,786 propositions. These components are evaluated separately to reflect the modular design of the framework. In particular, the summarization evaluation provides context for the input quality, whereas the fact-checking evaluation directly measures the effectiveness of the LLM-free verification module. Quantitative metrics, such as ROUGE, BERTScore, precision, recall, and F1-score, show that the framework produces logical narratives with minimal hallucinations.
The verification module independently identifies and evaluates such inconsistencies. Notably, factual consistency is assessed exclusively by this module and does not rely on any LLM-based judgment or scoring mechanism. The confusion matrix also shows that it can classify supported and unsupported claims well. These results show that the framework compares favorably to LLM-based evaluators, which was confirmed by other general benchmarks and reviews by clinicians.

\subsubsection{Training Loss Convergence}
\begin{figure}[]
    \centering
    \includegraphics[scale=0.55]{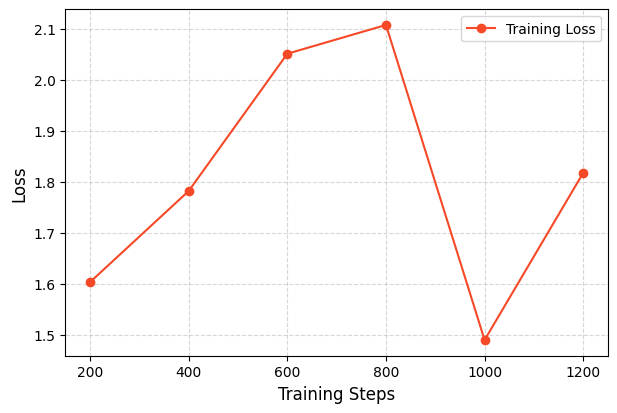}
    \caption{Training loss convergence over fine-tuning steps.}
    \label{fig:loss_curve}
\end{figure}

Figure~\ref{fig:loss_curve} shows the training loss curve over fine-tuning steps. The loss climbs up from approximately 1.60 at step 200 to a peak of 2.10 at step 800, then drops sharply to 1.49 at step 1000, and partially rebounds to 1.82 by step 1200. This indicates that the model is making significant weight updates during training instead of hitting a plateau early on. Such non-monotonic behavior is well-documented in fine-tuning with cosine learning-rate schedule, gradient accumulation, bf16 mixed precision, and gradient checkpointing \cite{johnson2023ps}. The long clinical sequence we are working with is also likely to play a role in this variability \cite{shirish2016large}. On the other hand, we used Unsloth for a parameter-efficient fine-tuning (LoRA) that can be responsible for variability in the reported loss at each step, especially when training on diverse clinical inputs \cite{hu2022lora}.

Figure~\ref{fig:loss_curve} shows the training loss curve over fine-tuning steps. The loss climbs up from approximately 1.60 at step 200 to a peak of 2.10 at step 800, then drops sharply to 1.49 at step 1000, and partially rebounds to 1.82 by step 1200. This indicates that the model is making significant weight updates during training instead of hitting a plateau early on. Such non-monotonic behavior is well-documented in fine-tuning with cosine learning-rate schedule, gradient accumulation, bf16 mixed precision, and gradient checkpointing \cite{johnson2023ps}. The long clinical sequence we are working with is also likely to play a role in this variability \cite{shirish2016large}. On the other hand, we used Unsloth for a parameter-efficient fine-tuning (LoRA) that can be responsible for variability in the reported loss at each step, especially when training on diverse clinical inputs \cite{hu2022lora}. It is also worth mentioning that Figure~\ref{fig:loss_curve} shows loss values at specific intervals, which can make local variations appear sharper than they are. Despite this trajectory, the final checkpoint loss remains within a competitive range with effective fine-tuning.

\subsection{LLM Summarization Performance}

\begin{figure}[]
    \centering
    \includegraphics[scale=0.4]{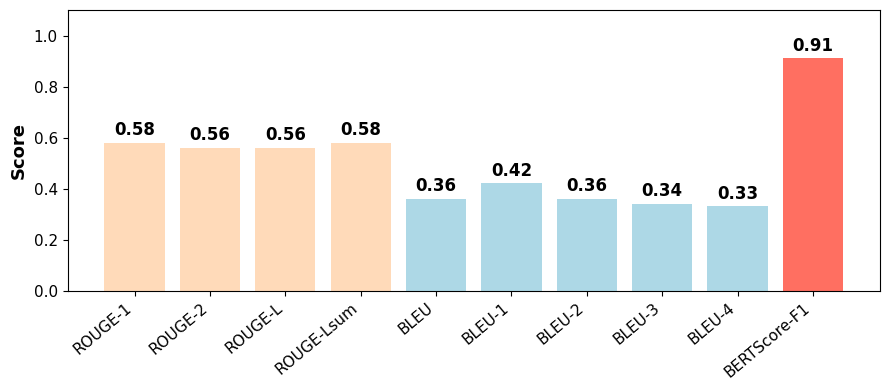}
    \caption{Performance metrics of the fine-tuned LLaMA-3.1-8B model on MIMIC-III discharge summaries.}
    \label{fig:results}
    \vspace{-10pt}    
\end{figure}

The fine-tuned LLaMA-3.1-8B model achieved strong performance by demonstrating its ability to generate clinically coherent summaries while minimizing hallucinations. Table \ref{tab:auto_metrics_compact} reports the evaluation metrics for the fine-tuned LLaMA-3.1-8B model. As shown in Figure~\ref{fig:results}, our fine-tuned model achieved strong performance with ROUGE-1, ROUGE-2, and ROUGE-L scores of 0.5797, 0.5580, and 0.5618, respectively, reflecting high lexical and structural alignment with the reference clinical summaries. A BLEU score of 0.3604 further indicates accurate n-gram overlap, while the high BERTScore (F1 = 0.9120) demonstrates strong semantic fidelity between the generated and reference texts. Collectively, these results indicate that the model produces concise and clinically relevant text.

To provide context for these values, we compare them to previously reported neural summarization systems on MIMIC-III and related clinical corpora. Prior LLM-based approaches, such as those by Tang et al. \cite{Tang2023LLMSummarization}, Xu et al. \cite{xu2024opportunities}, and Lin et al. \cite{LIN2025100868}, report ROUGE-L (or ROUGE-Lsum) and BERTScore values in a similar range when evaluated on long discharge summaries, indicating that our summarization performance is competitive with existing methods rather than an outlier. This suggests that the main contribution of our framework does not stem from unusually high ROUGE or BERTScore values alone, but from the downstream fact-checking module that explicitly controls factual consistency at the proposition level.

% The fine-tuned LLaMA-3.1-8B model achieved strong performance by demonstrating its ability to generate clinically coherent summaries while minimizing hallucinations.

% Figure~\ref{fig:results} reports the evaluation metrics for the fine-tuned LLaMA-3.1-8B model. 
% The model achieves strong summarization performance with ROUGE-1, ROUGE-2, and ROUGE-L scores of 0.5797, 0.5580, and 0.5618 respectively, reflecting high lexical and structural alignment with the reference clinical summaries. 
% A BLEU score of 0.3604 further indicates an accurate overlap of n-grams, while the high BERTS score (F1 = 0.9120) demonstrates strong semantic fidelity between the generated and reference texts. 
% Collectively, these results indicate that the model produces concise and clinically relevant text, and factual consistency is also ensured separately by the LLM-free fact-checking module.

 Overall, these results indicate that the LoRA-adapted LLaMA-3.1 model produces high-quality summaries whose residual factual errors can be systematically addressed by the verification pipeline. Propositions that remain incorrect after generation tend to be easily flagged by the proposition-level fact-checker. Combined with the logical verification module, the framework forms a reliable system for trustworthy clinical summarization.
% Overall, these results indicate that the LoRA-adapted LLaMA-3.1 model produces factually correct summaries. Those propositions that tend to be incorrect are easily flagged by the verification. Combined with the proposition-level verification pipeline, the framework forms a reliable system for trustworthy clinical summarization. 

\subsection{Fact Checker Performance}

The quantitative analysis in Table~\ref{tab:auto_metrics_compact} demonstrates that the proposed fact-checking system achieves strong overall performance in all essential metrics. The system can identify supported propositions with a precision of 0.8904 and an F1-score of 0.8556. On the 3,786-proposition test set, this corresponds to 2,340 true positives and only 288 false positives, i.e., a false discovery rate of 0.1096. In other words, when the system predicts that a proposition is supported, it is correct almost nine times out of ten, and the absolute number of false positives remains comparatively low. A recall value of 0.8234 reveals that most of the clinically valid facts in the summaries are found and verified with evidence in the EHR. The overall accuracy of 0.7913 on 3,786 propositions shows that the logical verification pipeline is strong. It also highlights its ability to address a wide range of factual errors, such as numerical, temporal, and logical errors. 

For evaluating the significance of each consistency check, we performed ablation analysis, as shown in Table \ref{tab:ablation_study}, where we progressively omitted one check at a time from our pipeline for verification purposes. These consistencies include negation, implication, temporal, mutual exclusiveness, numerical, and presence checks, which are carried out on a set of 3,786 propositions. The impact of omission is measured by changes in precision, recall, and F1-score for each omission, proving that all six checks are indeed important for the fact-checking module to be effective. 

The outcome of the ablation experiment, clearly demonstrates the importance of all six checks in the effectiveness of the fact-checking model. The removal of negation and number consistency checks leads to a significant decrease in precision and recall scores, with a drop exceeding 8\% in precision. The implication check and mutual exclusivity check also have an impact on the model's capacity to validate claims, leading to a 6-8\% decrease in precision and recall scores. The temporal consistency check and presence check do not have as much influence as the other four checks, but their absence still negatively impacts precision and recall.

These results collectively confirm the efficacy of the multi-layered fact-checking system in maintaining factual integrity in automatically generated clinical summaries. This step is crucial in order to consider LLMs as reliable systems for such critical sectors. Table \ref{tab:add_metrics_compact} reports additional evaluation metrics for the fact-checking system. Specificity (unsupported) equals 0.6949, which means that 69.49\% of actual unsupported claims are correctly identified. MCC equals 0.4866; this indicates that there is an overall moderate correlation between predictions and true labels. Balanced accuracy equals 0.7591, hence reflecting solid performance across both classes despite imbalance. Log loss equals 0.2288, which suggests well-calibrated probability predictions.

% The quantitative analysis in Table~\ref{tab:auto_metrics_compact} demonstrates that the proposed fact-checking system achieves a strong overall performance in all essential metrics. 
% The system can identify supported propositions with a precision of 0.8904 and an F1-score of 0.8556, retaining the number of false positives low. 
% A recall value of 0.8234 reveals that most of the clinically valid facts in the summaries are found and verified with evidence in the EHR. 
% The overall accuracy of 0.7913 on 3,786 propositions shows that the logical verification pipeline is strong. It also highlights its ability to address a wide range of factual errors, such as numerical, temporal, and logical errors. 
% These results collectively confirm the efficacy of the multi-layered fact-checking system in maintaining factual integrity in automatically generated clinical summaries. This step is crucial in order to consider LLMs as reliable systems for such critical sectors.

The findings reveal that our fact-checking approach proves to be quite efficient in discovering various types of contradictions, including numerical, temporal, and logical issues. The analysis suggests that this multi-layered verification strategy can identify both clear contradictions and the more nuanced inconsistencies that traditional methods might miss. The extra evaluation metrics in Table \ref{tab:add_metrics_compact} further support this observation, demonstrating a balanced performance across different classes. This suggests that merging semantic alignment with deterministic logical checks is especially powerful for ensuring reliability in critical areas like healthcare.

\begin{table}[ht!]
\centering
\scriptsize
\caption{Ablation study of consistency checks in the fact-checking module, showing the drastic impact on performance when each check is removed.} 
\label{tab:ablation_study}
\begin{tabular}{p{0.22\linewidth} p{0.45\linewidth}}
\hline
\textbf{Configuration} & \textbf{Performance Metrics} \\
\hline
\textbf{Full Model} & 6 checks (Negation, Implication, Temporal, \\
 & Mutual Exclusivity, Numerical, Presence) \\
\hline
\textbf{- Negation} & Precision: 0.8021, Recall: 0.7103, F1-score: 0.7521 \\
\textbf{- Implication} & Precision: 0.8103, Recall: 0.7217, F1-score: 0.7561 \\
\textbf{- Temporal} & Precision: 0.8256, Recall: 0.7404, F1-score: 0.7802 \\
\textbf{- Mutual Exclusivity} & Precision: 0.8038, Recall: 0.7142, F1-score: 0.7520 \\
\textbf{- Numerical} & Precision: 0.7469, Recall: 0.6558, F1-score: 0.6957 \\
\textbf{- Presence} & Precision: 0.7701, Recall: 0.6819, F1-score: 0.7236 \\
\textbf{Any 4 checks} & Precision: 0.7002, Recall: 0.6124, F1-score: 0.6529 \\
\textbf{Any 3 checks} & Precision: 0.6754, Recall: 0.5936, F1-score: 0.6315 \\
\hline
\end{tabular}
\end{table}

\subsection{Comparison with Existing Literature}
The presented method, called the Fact-Checking Module, outperforms previous methods and reaches the new state-of-the-art F1-score of 79.13\% on MIMIC-III \cite{johnson_mimiciii_2016}. This highlights the module’s superior performance in supporting clinical statements compared to current state-of-the-art models.

Previous systems, including Claude-1 \cite{anthropic2023a}, achieved an accuracy of 66.74\%, while its DOSSIER-extended version reached 70.53\%. Although Claude-2 \cite{anthropic2023b} demonstrated a performance of 70.65\% and Claude-2 (DOSSIER) achieved 78.62\%, our model exhibits a notable gain by exceeding 79\% without the need for large language models for verification. Models such as CodeLlama-13B \cite{roziere2023code}, MedAlpaca 7B \cite{han2025medalpacaopensourcecollection}, and ClinicalCamel 13B \cite{toma2023clinicalcamelopenexpertlevel} exhibited accuracies of 65.76\%, 46.74\%, and 32.51\%, respectively, indicating deficiencies in factual grounding and generalization. Domain-specialized architectures, including T5-EHRSQL \cite{lee2022ehrsql} and Asclepius 13B \cite{kweon2024publiclyshareableclinicallarge}, did not exceed 55\%, highlighting the performance gap.

\begin{table}[]
\centering
\scriptsize
\renewcommand{\arraystretch}{1.1}
\caption{Automatic Evaluation Metrics for Fact-Checking System.}
\label{tab:auto_metrics_compact}
\begin{tabular}{lcccc}
\toprule
\textbf{Precision} & \textbf{Recall} & \textbf{F1-Score} & \textbf{Accuracy} \\
\midrule
0.8904 & 0.8234 & 0.8556 & 0.7913 \\
\bottomrule
\end{tabular}
\end{table}

%_______________________________________________________
\begin{table}[]
\centering
\scriptsize
\renewcommand{\arraystretch}{1.2}
\caption{Additional Evaluation Metrics for Fact-Checking System.}
\label{tab:add_metrics_compact}
\begin{tabular}{lcccc}
\toprule
\textbf{Specificity (Unsupported)} & \textbf{MCC} & \textbf{Balanced Accuracy} & \textbf{Log Loss} & \textbf{FDR} \\
\midrule
0.6949 & 0.4866 & 0.7591 & 0.2288 & 0.1096 \\
\bottomrule
\end{tabular}
\end{table}
The experimental results validate the practical strength and efficacy of our consistent LLM-free verification pipeline. The proposed module ranks higher than the state-of-the-art counterpart with respect to factual consistency, transparency, and reproducibility. In Table ~\ref{tab:mimic_results}, we show an extensive comparison of our proposed method against state-of-the-art methods. Our proposed Fact-Checking Module achieves the highest accuracy (79.13\%), outperforms these baselines on MIMIC-III, including Claude-2 (DOSSIER) (78.62\%) and CodeLlama-13B (DOSSIER) (65.76\%). 
These findings validate the module’s capability for reliable, interpretable, and reproducible fact verification in clinical NLP.

\begin{table}[]
\scriptsize
\centering
\caption{Comparison of different models and their accuracy on the MIMIC-III clinical claim verification task.}
\begin{tabular}{l l c}
\toprule
\textbf{Model} & \textbf{Dataset} & \textbf{Accuracy (\%)} \\
\midrule
Claude-1~\cite{anthropic2023a} & MIMIC-III & 66.74 \\
Claude-1 (DOSSIER)~\cite{pmlr-v252-zhang24a} & MIMIC-III & 70.53 \\
CodeLlama-13B~\cite{roziere2023code} & MIMIC-III & 64.13 \\
CodeLlama-13B (DOSSIER)~\cite{pmlr-v252-zhang24a} & MIMIC-III & 65.76 \\
MedAlpaca 7B~\cite{han2025medalpacaopensourcecollection} & MIMIC-III & 46.74 \\
ClinicalCamel 13B~\cite{toma2023clinicalcamelopenexpertlevel} & MIMIC-III & 32.51 \\
Asclepius 13B~\cite{kweon2024publiclyshareableclinicallarge} & MIMIC-III & 39.26 \\
Llama2 7B 32k~\cite{touvron2023llama} & MIMIC-III & 32.81 \\
T5-EHRSQL~\cite{lee2022ehrsql} & MIMIC-III & 54.56 \\
Claude-2~\cite{anthropic2023b} & MIMIC-III & 70.65 \\
Claude-2 (DOSSIER)~\cite{pmlr-v252-zhang24a} & MIMIC-III & \underline{78.62} \\
\midrule
\textbf{Our Module} & MIMIC-III & \textbf{79.13} \\
\bottomrule
\end{tabular}
\label{tab:mimic_results}
\end{table}

\section{Discussion and Future Work}
Our proposed fact-checking system introduces a two-stage pipeline for clinical summarization. It combines a domain-specific LLaMA-3.1-8B generator, fine-tuned via Low-Rank Adaptation (LoRA) on MIMIC-III discharge summaries \cite{johnson2016mimic,hu2022lora}, which covers  broader categories such as respiratory, cardiovascular, infectious, metabolic, and neurological disorders etc , and it is robust to these diverse medical scenarios. The system is capable of creating coherent narratives, and also works as an independent fact-checking module that uses cosine similarity, numerical tests, and discrete logical checks for granular verification against Electronic Health Records (EHRs). The system recorded a ROUGE-1 score of 0.5797 and a BERTScore of 0.9120 for summary generation. The fact-checking module reached a precision of 0.8904, a recall of 0.8234, and an F1-score of 0.8556 over 3,786 propositions from 104 summaries. Clinician review (n = 2) judged roughly 85\% of a random subset of summaries as clinically acceptable, and the logical checks successfully highlighted the majority of remaining inconsistencies. The fact-checking module can also serve independently as a post-processing layer for any LLM-generated summary in the clinical domain, as it depends solely on deterministic logical checks rather than probabilistic language-model behavior \cite{vladika2023healthfc,thorne2018automated}.

Although the proposed system can also be applied to evaluate the factual accuracy of summaries generated by other LLMs, our fine-tuned LLaMA-3.1-8B model demonstrates superior summarization performance in this study. However, the current pipeline shows limited robustness in critical and less-explored medical domains where highly specialized knowledge is required. Examples include rare oncological subtypes (e.g., hematologic malignancies with complex staging), pediatric metabolic and genetic disorders, or transplant medicine and intensive care scenarios involving multi-organ failure. In such settings, the EHR may contain highly technical terminology, uncommon procedures, and nuanced temporal relationships that are under-represented in the training data and only partially covered by generic ontologies \cite{garcia2025check, joseph2024factpico}. This increases the risk that the generator omits key events or that the verifier fails to recognize domain-specific implications (for instance, subtle drug–drug interactions in oncology or transplant immunosuppression regimens). Robustness in these domains could be improved by integrating richer medical ontologies (e.g., subspecialty extensions of SNOMED-CT or disease-specific knowledge graphs), incorporating domain-specific lexicons, and fine-tuning components on curated data from specialized clinics \cite{tang2024minicheck, chen2025graphcheck}.

At the same time, the investigation and implementation of explicit reasoning mechanisms and automatic correction of unsupported outputs remain open in the current study. Our verifier currently functions as a binary gate: it labels propositions as Supported or Not Supported, but does not propose how to repair them or rewrite the summary. Future work could extend this by: (i) adding a symbolic reasoning layer that operates on the graph of extracted propositions (nodes as events, edges as temporal or logical relations) to propagate constraints and identify minimal sets of edits \cite{chen2025graphcheck}; (ii) generating candidate “corrected propositions” by substituting values or attributes from the EHR and feeding them back into a controlled rewriting step \cite{benabacha2024overview}; and (iii) incorporating clinician-in-the-loop workflows where flagged propositions are presented with explanations (“negation conflict”, “numerical mismatch”, “missing implied treatment”) and suggested fixes that the user can accept or modify \cite{kang2023evidencemap}. These directions would turn the module from a pure detector into an assistive tool that both diagnoses and helps correct factual errors, a need highlighted in recent clinical NLP hallucination surveys \cite{kim2025medical, ji2023survey}.

Future work will also concentrate on expanding the fact-checking module by integrating causal reasoning into the current temporal consistency assessments. Concretely, one direction is to model patient trajectories as causal or causal-inspired graphs, where nodes represent diagnoses, interventions, and outcomes, and edges encode plausible cause–and–effect relationships derived from clinical guidelines or learned from longitudinal EHR data \cite{chen2025graphcheck}. The temporal check could then be extended to verify not only that events occur in the correct order, but also that observed patterns are consistent with known causal pathways (for example, “initiation of anticoagulation should follow diagnosis of atrial fibrillation, not precede it”, or “improvement in oxygenation should not causally precede the start of mechanical ventilation”). Another strategy is to employ simple structural-causal models or counterfactual probes over the proposition graph, e.g., asking whether removing a key intervention would plausibly change downstream outcomes and flagging summaries that imply clinically implausible or causally inverted relationships \cite{goel2025zero, feng2025counterfactual}. These causal constraints can be combined with the existing numerical and temporal checks to increase the system’s ability to detect subtle, yet clinically important, hallucinations \cite{rawte2023survey,Huang_2025}.
\section{Conclusion}
This work introduced a two-stage framework for trustworthy clinical summarization that couples a LoRA-fine-tuned LLaMA-3.1-8B generator with an independent, LLM-free fact-checking module operating at the proposition level. The LoRA adaptation enables the base LLaMA-3.1-8B model to specialize on long, noisy discharge summaries from MIMIC-III while retaining its strong language modeling capabilities, resulting in summaries with competitive ROUGE and BERTScore metrics and clinically coherent narratives. Crucially, the generator is integrated with the verifier in a way that separates concerns: the LLaMA-3.1-8B component is optimized for fluent, clinically appropriate text, whereas the fact-checking module is optimized for fine-grained consistency with the underlying EHR. This division allows the system to achieve high overall performance, that is, precision of 0.8904 and F1-score of 0.8556 on 3,786 propositions, while maintaining transparency and reproducibility in the verification step.

At the same time, our results highlight several areas where robustness must be improved before deployment in safety-critical settings. The current extraction and verification pipeline performs well on common diagnoses and treatments but remains less reliable in highly specialized or under-represented domains, such as rare oncologic subtypes, pediatric metabolic disorders, or complex transplant cases. Addressing this will require augmenting the proposition schema and rule base with richer subspecialty ontologies, improving the coverage of entity normalization and implication rules, and incorporating causal and temporal reasoning mechanisms capable of capturing domain-specific treatment pathways. In addition, robustness could be strengthened through systematic evaluations on external datasets, cross-institutional validation, and uncertainty quantification that exposes when the verifier’s judgments are unreliable and should be escalated for human review.

Despite these limitations, the proposed framework has the potential to improve concrete clinical workflows. In a discharge-summary workflow, for example, the system could act as a post-hoc safety layer that automatically flags contradictions between the generated summary and the EHR, such as mismatched laboratory values, omitted comorbidities, or missing treatments implied by diagnoses before the clinician signs off. In medication reconciliation, it could check that the described therapies and dosages align with the structured medication list and allergy record, highlighting inconsistencies that might otherwise be overlooked. For inter-provider communication, the fact-checker could be applied to referral letters and handover notes to ensure that key events (e.g., procedures performed, complications, changes in code status) are accurately and consistently represented across documents. By embedding such proposition-level verification into routine documentation and handover processes, the framework can support safer, more reliable use of generative models in clinical practice and provide a principled foundation for future extensions that further integrate reasoning and correction capabilities.

\section*{Acknowledgments}
We express our heartfelt gratitude to Dr. Sajib Saha (MBBS, BCS, DO- National Institute of Ophthalmology, eleven years of experience) and Dr. Synthia Kor (MBBS, BCS, CMU, MO – Sylhet MAG Osmani Medical College, ten years of experience) for their valuable contribution as medical professionals to validate and verify our results. Both are licensed by the Bangladesh Medical \& Dental Council (BMDC).
\section*{Appendix}
\vspace{0.5em}
{\centering
\scriptsize
\renewcommand{\arraystretch}{1.2}
\begin{tabular}{c c c c}
\hline
\textbf{Short Form} &
  \textbf{Full Form} &
  \textbf{Short Form} &
  \textbf{Full Form} \\ \hline
\textbf{AI} &
  Artificial Intelligence &
  \textbf{LLMs} &
  Large Language Models \\ \hline
\textbf{AdamW} &
  Adaptive Moment Estimation with Weight Decay &
  \textbf{LOINC} &
  \begin{tabular}[c]{@{}c@{}}Logical Observation Identifiers\\  Names and Codes\end{tabular} \\ \hline
\textbf{AUC} &
  Area Under the Curve &
  \textbf{LoRA} &
  Low-Rank Adaptation \\ \hline
\textbf{BCS} &
  Bangladesh Civil Service &
  \textbf{MBBS} &
  Bachelor of Medicine, Bachelor of Surgery \\ \hline
\textbf{BERTScore} &
  BERT-based Similarity Score &
  \textbf{MCC} &
  Matthews Correlation Coefficient \\ \hline
\textbf{BioClinicalBERT} &
  Biomedical Clinical BERT &
  \textbf{MEDIQA-CORR} &
  \begin{tabular}[c]{@{}c@{}}Medical Error Detection and \\ Correction Shared Task\end{tabular} \\ \hline
\textbf{BLEU} &
  Bilingual Evaluation Understudy &
  \textbf{MIMIC-III} &
  Medical Information Mart for Intensive Care III \\ \hline
\textbf{BMDC} &
  Bangladesh Medical and Dental Council &
  \textbf{MIMIC-IV} &
  Medical Information Mart for Intensive Care IV \\ \hline
\textbf{CHECK} &
  \begin{tabular}[c]{@{}c@{}}Continuous Hallucination Elimination\\  and Knowledge Checking\end{tabular} &
  \textbf{MiniCheck} &
  Efficient Fact-checking Model \\ \hline
\textbf{CMU} &
  Chittagong Medical University &
  \textbf{NER} &
  Named Entity Recognition \\ \hline
\textbf{DO} &
  Doctor of Osteopathy &
  \textbf{NLI} &
  Natural Language Inference \\ \hline
\textbf{DOSSIER} &
  \begin{tabular}[c]{@{}c@{}}Dual-layer fact-checking framework\\  for Electronic Health Records\end{tabular} &
  \textbf{NLL} &
  Negative Log-Likelihood \\ \hline
\textbf{EHR} &
  Electronic Health Record &
  \textbf{NLP} &
  Natural Language Processing \\ \hline
\textbf{EHRs} &
  Electronic Health Records &
  \textbf{NSW} &
  New South Wales \\ \hline
\textbf{EMNLP} &
  Empirical Methods in Natural Language Processing &
  \textbf{NT} &
  Northern Territory \\ \hline
\textbf{FACTPICO} &
  Factuality Evaluation using PICO Framework &
  \textbf{PICO} &
  Population, Intervention, Comparator, Outcome \\ \hline
\textbf{FactSelfCheck} &
  Fact-level Self Checking &
  \textbf{PMLR} &
  Proceedings of Machine Learning Research \\ \hline
\textbf{FDR} &
  False Discovery Rate &
  \textbf{RAG} &
  Retrieval-Augmented Generation \\ \hline
\textbf{Finch-Zk} &
  Zero-knowledge hallucination detection framework &
  \textbf{ROUGE} &
  Recall-Oriented Understudy for Gisting Evaluation \\ \hline
\textbf{GPU} &
  Graphics Processing Unit &
  \textbf{RxNorm} &
  \begin{tabular}[c]{@{}c@{}}Normalized Naming System for\\  Generic and Branded Drugs\end{tabular} \\ \hline
\textbf{GraphCheck} &
  Graph-based Fact-checking Model &
  \textbf{SNOMED-CT} &
  \begin{tabular}[c]{@{}c@{}}Systematized Nomenclature \\ of Medicine Clinical Terms\end{tabular} \\ \hline
\textbf{ICD-9} &
  \begin{tabular}[c]{@{}c@{}}International Classification of\\  Diseases, Ninth Revision\end{tabular} &
  \textbf{Sentence-BERT} &
  \begin{tabular}[c]{@{}c@{}}Sentence Bidirectional Encoder \\ Representations from Transformers\end{tabular} \\ \hline
\textbf{ICLR} &
  International Conference on Learning Representations &
  \textbf{USMLE} &
  United States Medical Licensing Examination \\ \hline
\textbf{LLM} &
  Large Language Model &
   &
   \\ \hline
\end{tabular}
\par}
\vspace{1em}

\section*{Statements and Declarations}
\textbf{Funding:} This study does not include any external funding.

\textbf{Ethical Approval:} Not applicable

\textbf{Competing interests:} The authors state that they have no known financial conflicts of interest or personal relationships that could have influenced the work presented in this paper.

\textbf{Data Availability:} In this work, we utilized the publicly available MIMIC-III dataset  \cite{johnson2016mimic}.

\textbf{CRediT authorship contribution statement:}

\textbf{Musarrat Zeba}: Conceptualization, Literature review, Formal analysis, Methodology, Writing: original draft, Validation, Writing – review \& editing;
\textbf{Abdullah Al Mamun}: Conceptualization, Literature review, Formal analysis, Methodology, Writing: original draft, Validation, Writing – review \& editing; 
\textbf{Kishoar Jahan Tithee}: Literature review, Formal analysis, Methodology, Writing: original draft, Validation, Writing – review \& editing; 
\textbf{Debopom Sutradhar}: Formal analysis, Supervision, Validation, Writing – original draft, Writing – review \& editing; 
\textbf{Mohaimenul Azam Khan Raiaan}: Conceptualization, Literature review, Formal analysis, Supervision, Project administration, Writing – original draft, Writing – review \& editing; 
\textbf{Saddam Mukta}: Validation, Formal analysis, Writing – review \& editing;
\textbf{Reem E. Mohamed}: Validation, Formal analysis, Writing – review \& editing;
\textbf{Md Rafiqul Islam}: Validation, Formal analysis, Writing – review \& editing;
\textbf{Yakub Sebastian}: Validation, Formal analysis, Writing – review \& editing;
\textbf{Mukhtar Hussain}: Validation, Formal analysis, Writing – review \& editing;
\textbf{Sami Azam}: Conceptualization, Supervision, Project administration, Validation, Writing: review.

\end{document}